\documentclass[sn-mathphys,Numbered,iicol]{sn-jnl}

\usepackage{graphicx}%
\usepackage{multirow}%
\usepackage{amsmath,amssymb,amsfonts}%
\usepackage{amsthm}%
\usepackage{mathrsfs}%
\usepackage[title]{appendix}%
\usepackage{xcolor}%
\usepackage{textcomp}%
\usepackage{manyfoot}%
\usepackage{booktabs}%
\usepackage{algorithm}%
\usepackage{algorithmicx}%
\usepackage{algpseudocode}%
\usepackage{listings}%
\usepackage[caption=false,font=scriptsize,labelfont=sf,textfont=sf]{subfig}
\usepackage{stfloats}

\begin{document}

\title[LiLO: Lightweight and low-bias LiDAR Odometry method based on spherical range image filtering]{LiLO: Lightweight and low-bias LiDAR Odometry method based on spherical range image filtering}


\author{\;\;\;\;Edison P. Velasco-Sánchez, Miguel Ángel Muñoz-Bañón, \\ Francisco A. Candelas, Santiago T. Puente and Fernando Torres. }

\affil{The authors are with the Group of Automation, Robotics and Computer Vision, \orgname{AUROVA}, University of Alicante, \city{San Vicente del Raspeig}, \postcode{03690}, \state{Alicante}, \country{Spain}}

\abstract{In unstructured outdoor environments, robotics requires accurate and efficient odometry with low computational time. Existing low-bias LiDAR odometry methods are often computationally expensive. To address this problem, we present a lightweight LiDAR odometry method that converts unorganized point cloud data into a spherical range image (SRI) and filters out surface, edge, and ground features in the image plane. This substantially reduces computation time and the required features for odometry estimation in LOAM-based algorithms. Our odometry estimation method does not rely on global maps or loop closure algorithms, which further reduces computational costs. Experimental results generate a translation and rotation error of 0.86\% and 0.0036°/m on the KITTI dataset with an average runtime of 78ms. In addition, we tested the method with our data, obtaining an average closed-loop error of 0.8m and a runtime of 27ms over eight loops covering 3.5Km. The code is available at: \href{https://github.com/AUROVA-LAB/applications}{\tt \small https://github.com/AUROVA-LAB/applications.}}




\maketitle

\section{Introduction}\label{sec1:introduction}

An accurate odometry system is one of the requirements for a robust autonomous driving system. Robotic odometry systems can be generated by using or fusing different types of sensors such as encoders, cameras, GPS and IMU (Inertial Measurement Unit) systems \cite{BLUE2020deeper,zhao2021accurate,alkendi2021state}. Depending on the robustness of the system, they tend to have high or low odometry drift as the robotic system advances along a trajectory. Therefore, markers or key-points are necessary for robotic localization to increase odometry robustness and decrease cumulative drift. These key-points are generated from monocular images, stereo images or point clouds. LiDAR (Light Detection And Ranging) devices, which generate a three-dimensional (3D) point cloud, are resistant to illumination variations and weather changes, making them ideal for outdoor environments \cite{debeunne2020review}. Due to the advantages of LiDAR sensors, odometry systems using point cloud data have become popular in recent years. LiDAR-based odometry can be generated by matching a point cloud to a local 3D map \cite{zhang2014loam}. In general, these methods use an Iterative Closest Point (ICP)-based algorithm, which estimates the transformation between two point clouds by minimizing their translation and rotation iteratively \cite{besl1992method, dellenbach2021s}. However, a large number of points increases the computational cost. To reduce processing costs, \mbox{KD-Tree} indexing methods are often used, which decreases the computation time with a distance threshold to limit the search \cite{reviewKDTREE,NICP,LiTAMIN}.  The most popular system for point cloud odometry is LiDAR Odometry And Mapping (LOAM) \cite{zhang2014loam}. This method obtains a low trajectory bias by implementing two algorithms. The first one, runs the odometry at high frequency, but with low fidelity. And the second one achieves fine point cloud matching and registration at lower frequency by performing point-to-edge and point-to-plane ICP matching by extracting corner and surface features from a cloud of input points. This feature extraction is a procedure that has been optimized in several research works \cite{zhang2017loam,legoloam2018,wang2021floam,garcia2022liodom}. The features are extracted from the raw point cloud and depends on the smoothness of the points on the local surface, where a maximum value is considered to be an edge and minimum values are surfaces. These LiDAR odometry systems are often computationally expensive and are challenging to integrate as stages in other robotic applications, such as path planning, motion control, precise positioning, complex localization systems \cite{slam} or geo-localization methods \cite{icra-data-association, data-association}. For this reason, there is a constant search for lightweight, low-bias, computationally efficient odometry systems that can be integrated into more complex stages of mobile robotics.


\begin{figure*}[htbp]
    \centering
    \includegraphics[width=0.95\textwidth]{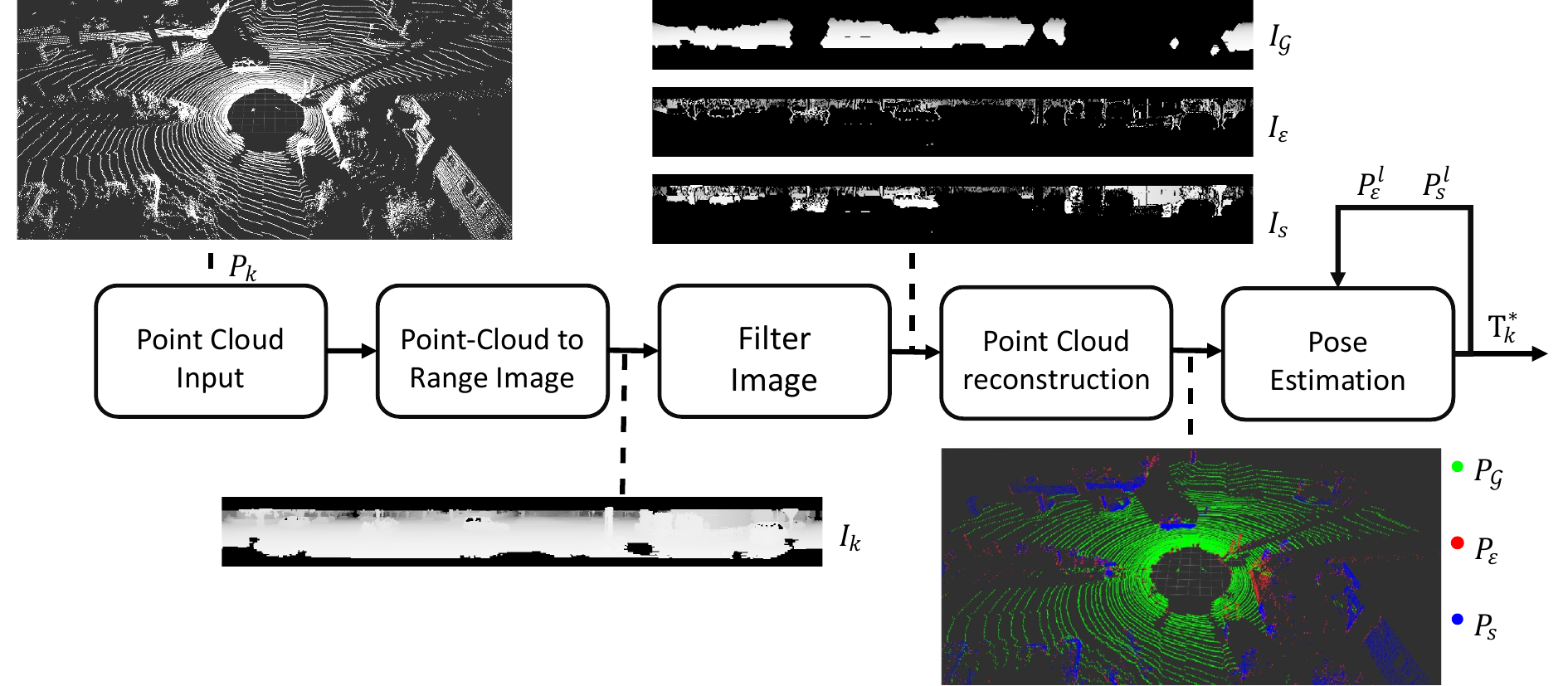}
    \caption{Pipeline of our LiLO method for odometry estimation, which filters the ground
    $\mathbf{I}_{\mathcal{G}}$, edges $\mathbf{I}_{\mathcal{E}}$ and surfaces $\mathbf{I}_{\mathcal{S}}$ in a SRI obtained from a point cloud. Point cloud reconstruction: ground $\mathbf{P}_{\mathcal{G}}$(green), edge $\mathbf{P}_{\mathcal{E}}$(red) and surface $\mathbf{P}_{\mathcal{S}}$(blue). To estimate the pose $\mathbf{T^*_k}$ we use the local feature maps $\mathbf{P}_\mathcal{E}$ and $\mathbf{P}_\mathcal{S}$ which possess the edge and surface features of the scenario respectively.}
    \label{fig:pipeline_LOFRI}
\end{figure*}
In this article, we introduce LiLO (Lightweight LiDAR Odometry) Fig.\ref{fig:pipeline_LOFRI}, a LiDAR odometry method with a lightweight but efficient pre-processing of ground, edge and surface feature extraction from a point cloud by converting them into SRI and gradient image filtering with the Sobel operator for 64-channel LiDAR, as well as image filtering with Fourier transform masks for 16-channel LiDAR. These filtered features in the SRI are converted back to a point cloud to reduce the density of the point cloud by voxel filtering. In order to make our method even lighter, it does not generate a global map of the scene, instead it uses the frame to local map technique, which, in order to decrease the computational cost, generates a local map using the \mbox{KD-Tree} indexing method, and also cleans this map continuously as the robot translation increases. Furthermore, our LiDAR odometry method does not require additional external sensors, e.g. inertial measurement unit, geo-localization sensors and/or cameras. 

The presented method of feature filtering and pose estimation with local mapping is lightweight and maintains good results when compared to versions of LOAM that also do not use global mapping. Our research achieves an odometry system that, despite being lightweight, maintains a low-bias. This system has been tested as an odometry stage in complete localization \cite{icra-data-association, data-association} and path planning \cite{OSM-navigation} methods without loss of accuracy. In addition, our approach was tested and has a high performance to be used as an odometry stage for precise positioning on a mobile robotic platform in waste collection in unstructured outdoor environments \cite{blue2022manipulacion}.

This document is organized as follows: in Section 2, we introduce the related work in point cloud odometry.  In Section 3, we explain the operation of the proposed method, where we describe how to estimate the position of a LiDAR system by obtaining the features of the point clouds by converting them into SRI. Section 4 shows the results of experiments with the KITTI dataset where we compare our technique with several state-of-the-art methods, and on a UGV platform that has a 16-beam LiDAR sensor. In addition, we analyze which is the appropriate resolution of conversion from point cloud to SRI and which is the combination of feature group (edges, surfaces and ground) that our method has the lowest translation and rotation error. Finally, Section 5 presents the conclusions of this article and suggests future research lines.

\section{Related work}

\subsection{LiDAR Odometry and Mapping}
\label{section:LOAM}
LOAM is one of the most explored methods in odometry based on ICP matching and several papers and applications have been derived from it, such as the following. The F-LOAM method \cite{wang2021floam} uses motion prediction and structure cost optimization for location. The LeGO-LOAM algorithm \cite{legoloam2018}, uses the presence of a ground plane in its segmentation and optimization steps by means of a Levenberg-Marquardt based optimization method to achieve pose estimation. The SuMa algorithm \cite{behley2018efficient} uses surface features to generate maps and performs frame-to-model ICP using point-to-plane residuals. Another method based on LOAM is CT-ICP \cite{CT-ICP} which uses the elastic distortion of the scan during point cloud registration to increase accuracy, thus is robust to high frequency and discontinuity movements. IMLS-SALM \cite{IMLS-SLAM}, on the other hand proposes a sampling strategy based on LiDAR scans, then defines a model of the scenario with the LiDAR scans previously located and uses the Implicit Moving Least Squares (IMLS) surface representation.
LiTAMIN2 proposes an ICP method using symmetric KL divergence to significantly improve the speed of LiDAR-based SLAM, achieving tracking and mapping with 500-1000 Hz processing. LIODOM \cite{garcia2022liodom} presents a LOAM-based method that represents the environment through a data structure combining a hash scheme, allowing quick access to any section of the map. 
Other approaches use deep learning for pose estimation, such as PWCLO-Net \cite{PWCLO}, this uses the Pyramid, Warping, and Cost volume (PWC) structure for the LiDAR odometry task is constructed to refine the estimated pose in a hierarchical approach from coarse to fine. In \cite{nubert2021self} introduces a work based on self-supervised learning-based for robot pose estimation directly from LiDAR data, this method does not require any labeling for training.

Systems that use the LOAM method with other sensors are presented in \cite{GR-LOAM, LIO-SAM, koide2022globally}, where the robot motion is estimated by the fusion of LiDAR and the IMU. DV-LOAM \cite{DV-LOAM} uses a direct visual LiDAR odometry and mapping framework that combines a monocular camera with sparse and accurate LiDAR range measurements. V-LOAM \cite{zhang2015visual}, uses visual odometry for ego-motion estimation and LiDAR-based odometry for refinement. T-LOAM \cite{T-LOAM} truncated least squares algorithm is employed, and proposes a multi-region ground extraction method and a dynamic curved voxel clustering to achieve 3D point cloud segmentation, where edge, surface, sphere and ground features are discretized. SLOAM \cite{S-LOAM} proposes a pose optimization based on semantic LIDAR features by simultaneously obtaining tree models while estimating the pose of a robot in a forest. 

LiLO is presented as a method based on the LOAM pose estimation method and the extraction of features from the point cloud by filtering them in the image plane with gradient filtering methods. This approach makes the filtering of point cloud features less computationally complex and also scalable depending on the resolution of the image plane. Our method does not generate global maps or loop closing to make it lightweight so that it can be implemented in more complex localization algorithms. Furthermore, based on the implementation of our method, the computational cost can be decreased by modifying the resolution of the projected SRI of the point cloud. This could increase the odometry bias, but working in conjunction with complex localization approaches \cite{icra-data-association, data-association} can achieve robust results.

\subsection{Spherical Range Image Odometry}
The methods based on SRI use 3D point clouds converted into a 2D planar image that represents the scenery. For example, DiLO \cite{Dilo} is a direct LIDAR-based odometry process based on a SRI, which uses an SRI normal as texture information and estimates frame-to-frame positioning by obtaining features in consecutive images. In the work presented in \cite{ELI}, both the non-terrestrial SRI and the bird's eye view terrestrial map are considered, and a robust adaptive range normal estimation method is used for LiDAR scan registration.

Other approaches propose to work with SRI and machine learning. This is the case of the method DMLO \cite{dmlo}, which is a deep learning approach to extract high-confidence matched pairs from successive LiDAR scans working with SRI. In contrast, LodoNet \cite{lodonet}, uses a convolutional neural network architecture with SRI to infer rotation and translation information from pairs of matched key-points (MKPs) extracted from consecutive scans. 

These papers show that features can be extracted from a 3D scenario by converting LiDAR sensor data into a SRI and then used for pose estimation. The simplicity of applying filtering methods on the image plane reduces problem complexity and feature extraction time. Our method filters scenery features from a SRI by applying image filters with the Sobel method or the Fourier mask filtering method, both computationally lightweight and adjustable methods for any type of image resolution. Thus, the same methodology can be used for feature extraction from LiDAR sensors with different number of laser beam layers.

\section{FILTERED SPHERICAL RANGE IMAGE}
\subsection{LiDAR point cloud to Spherical Range Image}

\begin{figure}[htbp]
    \centering
    \includegraphics[width=6cm]{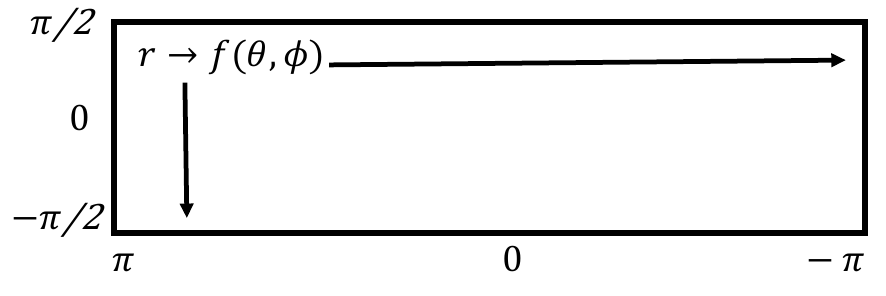}
    \caption{Transformation of point clouds in $\mathbb{R}^3$ into a SRI $\mathbf{I}_{SRI}$ in $\mathbb{R}^2$.}
    \label{fig:range_image_representation}
\end{figure}

The point cloud filtering method for obtaining edge, surface and ground features requires a conversion of the input point cloud into a SRI. This converts the 3D $\mathbf{P}_k$ data \eqref{eqn:PointCloud representation} into a planar spherical projection as shown in Fig. \ref{fig:range_image_representation}. The array generated by the range data $\mathbf{r}_k\geq0$ extending horizontally at $-\pi<\mathbf{\theta}_{k}<\pi$ and vertically at $-\frac{\pi}{2}<\mathbf{\phi}_{k}<\frac{\pi}{2}$ is called the SRI $\mathbf{I}_{SRI}$, which is described in (\ref{eqn:SRI_representation}).


\begin{equation}
    \mathbf{P}_k= [x_k,y_k,z_k] \newline
    \label{eqn:PointCloud representation}
\end{equation}

\begin{equation}
\begin{matrix}
    & \;\;\;\;\;\mathbf{r}_k = ||x_k+y_k+z_k||_2\\
    \;\;\;\;\mathbf{I}_{SRI} = \emph{f}(\mathbf{r}_k,\mathbf{\theta}_k,\mathbf{\phi}_k) & \!\!\mathbf{\theta}_k= \arctan  \left (  \frac{y_k}{z_k} \right )\\
    & \!\!\mathbf{\phi}_k= \arcsin \left ( \frac{z_k}{r_k} \right )\\
    \label{eqn:SRI_representation}
\end{matrix}
\end{equation}

$\mathbf{I}_{SRI}$ is transformed to a matrix $\mathbf{I}$ as shown in \eqref{eq:I rangeImageArray}. This matrix has a width index $j$ \eqref{eq:I range_m} and height index $i$ \eqref{eq:I range_n} calculated by shifting the original angles $\mathbf{\theta}$ and $\mathbf{\phi}$ to the range $0\!<\!\mathbf{\theta'}\!<\!360$ and $0\!<\!\mathbf{\phi'}\!<\!n_b$. Where $n_b$ is the number of beams laser sensor, and the values $X_{res}$ and $Y_{res}$ correspond to the desired image resolution. In the experimentation with a 16-beam laser sensor, an bilinear interpolation is implemented in the SRI using a 2D interpolation method \cite{kirkland2010bilinear} and developed in \cite{EPVelasco_lidar}.

\begin{equation}
    \mathbf{I}_{(i,j)} = \begin{bmatrix}
     r_{(0,0)}  &\cdots  & r_{(N-1,0)}\\ 
     \vdots   &  \ddots& \vdots  \\  
     r_{(0,M-1)}   &   \cdots&  r_{(N-1,M-1)}  
   \end{bmatrix}
    \label{eq:I rangeImageArray}
\end{equation}   

\begin{equation}
    i = \mathbf{\phi'} \cdot Y_{res}
    \label{eq:I range_n}
\end{equation}
\begin{equation}
    j = \mathbf{\theta'} \cdot X_{res}
    \label{eq:I range_m}
\end{equation}

The transformation $\mathbf{I}_{SRI}\!\rightarrow\!\mathbf{I}$ generates a matrix $\mathbf{I}$ with dimensions $MxN$, which contains all the distances $\mathbf{r}_k$ of each $k$ in the point cloud displayed horizontally at $j$ and vertically $i$. In order to represent the results of $\mathbf{I}$ as an image, the distances values are normalized to a gray-scale image.

\subsection{Image Filtering}
Point-to-edge and point-to-plane data matching work with feature sets representing the edges and surfaces of each scene in the point cloud. In contrast to LOAM-based methods (Section \ref{section:LOAM}), where edges and surfaces are differentiated with a smoothness parameter, we first need to apply filters on the SRI to determine those parameters. To filter the features in the point cloud, we consider that in the SRI representation, the points corresponding to the ground of the scenario are lines with a similar intensity value along the horizontal axis. Furthermore, we assume that the vertical line features of the SRI represent the edges, and the rest features are surfaces of the point cloud.

\subsubsection{Features segmentation}
\label{section:Edge and surface segmentation}
The Sobel operator represents the discrete derivatives that compute an approximation of the gradient of an image intensity function. This is established by convolving the original image $\mathbf{I}$ with a mask, as shown in \eqref{eqn:edge_masks} and \eqref{eqn:ground_masks}. Thus, with the selected masks, the vertical and horizontal characteristics of the original SRI are filtered out.

\begin{equation}
        \mathbf{M}_{\mathcal{E}} = \scriptsize { \begin{bmatrix}
             -1&0&1 \\
             -2&0&2\\
             -1&0&1
        \end{bmatrix}} *  \normalsize{\mathbf{I}}
    \label{eqn:edge_masks} 
\end{equation}  

\begin{equation}
        \mathbf{M}_{\mathcal{G}} = \scriptsize { \begin{bmatrix}
             -1&-2 &-1\\ 
             ~0&~0 &~0\\ 
             ~1&~2 &~1
        \end{bmatrix}} *  \normalsize{\mathbf{I}}
    \label{eqn:ground_masks} 
\end{equation}

These masks are applied with an element-to-element product $\odot$ to the original SRI. The $\mathbf{I}_{\mathcal{E}}$ edge \eqref{eq:Ie compute}  and $\mathbf{I}_{\mathcal{G}}$ ground features \eqref{eq:Ig compute} are extracted with the $\mathbf{M}_{\mathcal{E}}$ and $\mathbf{M}_{\mathcal{G}}$ mask respectively. Finally, the characteristics considered as surfaces of the spherical  range image, and represented by $\mathbf{I}_{\mathcal{S}}$ surface array \eqref{eq:Is compute} is obtained by subtracting the $\mathbf{I}_{\mathcal{E}}$ and the $\mathbf{I}_{\mathcal{G}}$ array from the original image $\mathbf{I}$.

\begin{equation}
    \mathbf{I}_{\mathcal{E}} = \mathbf{I}\odot M_{\mathcal{E}}
    \label{eq:Ie compute}
\end{equation}   
\begin{equation}
    \mathbf{I}_{\mathcal{G}} = \mathbf{I}\odot M_{\mathcal{G}}
    \label{eq:Ig compute}
\end{equation}
\begin{equation}
    \mathbf{I}_{\mathcal{S}} = \mathbf{I}-(\mathbf{I}_{\mathcal{E}}+\mathbf{I}_{\mathcal{G}})
   \label{eq:Is compute}
\end{equation}

\begin{figure*}[!t]
    \centering
    \includegraphics[width=1.0\textwidth]{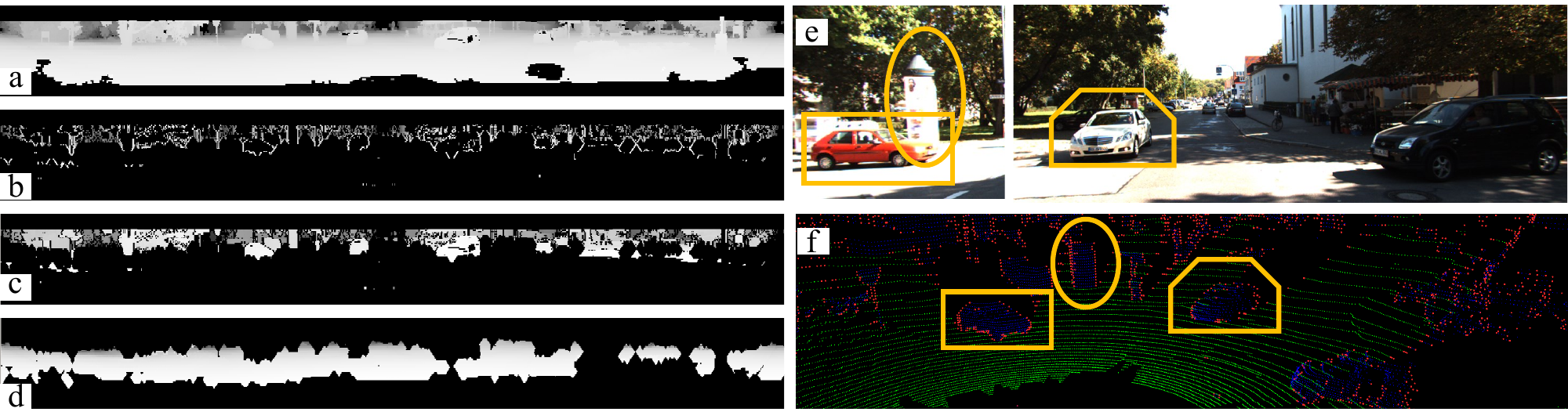}
    \caption{Segmentation of the point cloud of sequence 08 of the KITTI dataset applying the ground segmentation method of Section \ref{section:Edge and surface segmentation}. \textbf{(a)} SRI of of the point cloud of sequence 08. \textbf{(b)} Edge features of the spherical  range image. \textbf{(c)} Surface features of the SRI. \textbf{(d)} Ground feature of the spherical  range image.  \textbf{(e)} Photographs 5 and 34 of sequence 08 of the KITTI dataset. \textbf{(f)} Segmented features of the point cloud. Red, blue and green points are edges, surfaces and ground respectively.}
    \label{fig:rangeImage_sobel}
\end{figure*}

Fig. \ref{fig:rangeImage_sobel} shows the feature segmentation with the Sobel operator of the sequence 08 of the KITTI dataset compared to the RGB images of the corresponding scenario.

In this way, we filter the features of a point cloud in the image plane by convolution masks with the SRI.  Thus, feature extraction is computationally lightweight and can be scaled to different types of SRI resolutions, further reducing the computational cost but compromising the pose estimation accuracy by increasing the drift error. Which makes the feature extraction method accurate and lightweight according to the application of the method in robotics.

\subsubsection{Filtering in the frequency domain}
\label{section:Filtering in the frequency domain}
Another method of SRI filtering representing ground features that we propose is by frequency domain filtering. The point cloud is projected in a cylindrical shape due to the mechanical characteristics of the LiDAR sensor, so the points projected on the ground have a circular shape. This set of points with circular shape is represented in the SRI as horizontal lines \cite{badino2011fast}, which in the frequency domain are represented as features with a frequency close to zero. Using the frequency domain of a function allows us to analyze the frequency characteristics of an image. The fast Fourier transform $\mathcal{F}$ is used to find the frequency domain of a discrete process. The resulting image is a frequency domain representation of the original spherical  range image. Each $\mathbf{\hat{I}}$ point represents a particular frequency contained in the original $\mathbf{I}$ image. Using OpenCV library \cite{opencv_library} we transform a gray-scale image to a frequency domain image. The Fourier transform process of the spherical  range image is represented in \eqref{eqn:Fourier}.

 \begin{equation}
      \mathbf{\hat{I}}_{(i,j)} \overset{\mathcal{F}}{\rightarrow} \mathbf{I}_{(i,j)}
\label{eqn:Fourier}
\end{equation}

\begin{figure}[htbp]
    \centering
    \includegraphics[width=1.0\columnwidth]{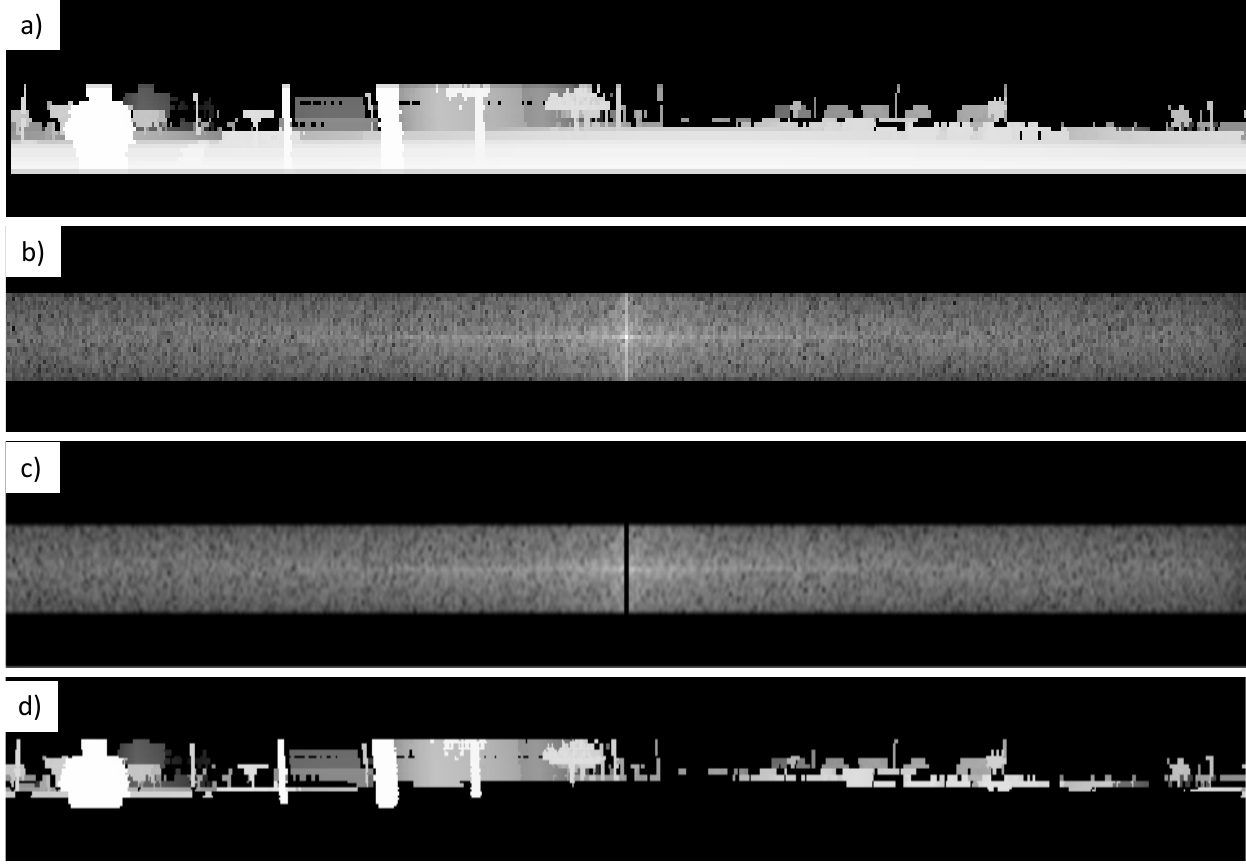}
    \caption{SRI from point cloud of Velodyne VLP-16 LiDAR sensor and ground point removing process by a frequency domain filtering. \textbf{(a)} SRI $\mathbf{I}$ representation of the point cloud $\mathbf{P}_k$. \textbf{(b)} Fourier transformed image $\mathbf{\hat{I}}$ representation of the original image $\mathbf{I}$. \textbf{(c)} Mask $ \mathbf{M}_{\mathcal{F}}$ applied to the Fourier transformed image $\mathbf{\hat{I}}$. \textbf{(d)} Filtered image $\mathbf{I'}$ obtained as the inverse Fourier transform of $\mathbf{\hat{I}}$ after applying the $\mathbf{M}_{\mathcal{F}}$ mask.}
    \label{fig:rangeImage_fourier}
\end{figure} 

The zero frequency component is placed in the upper-left corner of the resulting image. For a better interpretation of the results, the low frequencies are placed in the center of the image (Fig. \ref{fig:rangeImage_fourier}b). In this way, the lowest frequencies corresponding in $\mathbb{R}^3$ to the ground points can be filtered out using the Fourier mask $\mathbf{M}_{\mathcal{F}}$ represented in (\ref{eq:fourierMask}). 

\begin{equation}  
    \mathbf{M}_{\mathcal{F}(i,j)} = \left\lbrace
        \begin{array}{ll}
         0,   & \left (\frac{N}{2}-1\right  ) < j < \left (\frac{N}{2}+1\right  )\\
         1,   & otherwise
        \end{array}
        \right.\\
        \label{eq:fourierMask}
\end{equation}

Where $N$ is the width of the image, in this way we generate a low-pass filter that removes the low frequencies from the SRI, as shown Fig. \ref{fig:rangeImage_fourier}c. 

Hence, the filtered SRI $\mathbf{I'}$ \eqref{eq:FourierInverse} is the result of the inverse Fourier transform $\mathcal{F}\;^{-1}$ of the original image in the frequency domain $\mathbf{\hat{I}}$ \eqref{eqn:Fourier}  applying the mask $\mathbf{M}_{\mathcal{F}}$ \eqref{eq:fourierMask} by means of a element-to-element product. Therefore, a SRI $\mathbf{I'}$ is obtained with the ground characteristic points removed, as shown in Fig. \ref{fig:rangeImage_fourier}d. 

\begin{equation}  
        \mathbf{\hat{I}} \odot  \mathbf{M}_{\mathcal{F}}   \overset{\mathcal{F}\;^{-1}}{\rightarrow} \mathbf{I'} 
\label{eq:FourierInverse}
\end{equation}

In the experimentation we analyze which method, Sobel with mask \eqref{eqn:ground_masks} or frequency domain filtering, is successful in extracting ground features in the SRI. In this way, for the point cloud reconstruction, we compare the SRI of ground segmentation obtained with equations \eqref{eq:Ig compute} or \eqref{eq:FourierInverse} as ground features.  

\subsection{Point cloud reconstruction}
\label{pcl_reconstruction}
The proposed LiLO odometry method performs a point-to-egde data matching with a dataset in $\mathbb{R}^3$ in order to estimate the pose. The transformation of a SRI to a 3D point cloud $\mathbb{R}^2\!\rightarrow\!\mathbb{R}^3$ calculates the components  $\{x, y, z\}$ based on a 2D image that has depth information, referred to as a 2.5D image. The components $\mathbf{x}_{\mathcal{E}}, \mathbf{x}_{\mathcal{S}}, \mathbf{x}_{\mathcal{G}}$ and $\mathbf{y}_{\mathcal{E}}, \mathbf{y}_{\mathcal{S}}, \mathbf{y}_{\mathcal{G}}$ correspond to the $x$ \eqref{eq:x_reconstruction} and $y$ \eqref{eq:y_reconstruction} components of the edge, surface and ground features, are calculated from the cosine and sine function of the azimuth angle $\mathbf{\omega}$  and the modulus of the corresponding SRI ($\mathbf{I}_\mathcal{E}$, $\mathbf{I}_\mathcal{S}$ and $\mathbf{I}_\mathcal{G}$) with the $\mathbf{z}$ component of each feature of SRI. To obtain this modulus, we normalize back the features of SRI to the range values of the point cloud $\mathbf{P}_k$. 
\begin{equation}
    \mathbf{x} = ||\mathbf{I}-\mathbf{z}||_2\cdot \cos(\mathbf{\omega}) 
        \label{eq:x_reconstruction}
\end{equation}
\begin{equation}    
    \mathbf{y} = ||\mathbf{I}-\mathbf{z}||_2\cdot \sin(\mathbf{\omega}) \\
    \label{eq:y_reconstruction}
\end{equation}

 This $\mathbf{\omega}$ value is calculated by the normalized column number $j$ between $\pi$ to $-\pi$ and the width of the image $N$, as equation (\ref{eq:CalculateAngular_3dRecosntruction}) describe.
 
\begin{equation}
    \mathbf{\omega}_{(i,j)}=\left [\pi-\frac{(2\pi \cdot j)}{N} \right ]
    \label{eq:CalculateAngular_3dRecosntruction}
\end{equation}

The $\mathbf{z}$ \eqref{eq:Zesg compute} components of each point cloud feature are determined as all elements of $\mathbf{z'}$ \eqref{eq:Zesg compute} where their corresponding $\mathbf{x}, \mathbf{y}$ component is different to zero respectively.
\begin{equation}
   \mathbf{z} \subset \mathbf{{z}'} \; \forall \; (\mathbf{x} \wedge \mathbf{y}) \ne 0 
    \label{eq:Zesg compute}
\end{equation}
To calculate $\mathbf{z'}$, we first convert the $z$ components of all points of original point cloud $\mathbf{P}_k$ into a matrix of the same dimensions and shape of the SRI $\mathbf{I}$ defined in \eqref{eq:I rangeImageArray}. Thus, we denote as $\mathbf{Pz}$ the $z$ components of $\mathbf{P}_k$ matrix. In this way, $\mathbf{z'}$ is calculated as all values of $\mathbf{Pz}$, where the coordinates $(i,j)$ are all the integer elements of the normalization of the angles  $\mathbf{\theta'}$ and $\mathbf{\phi'}$ with the resolution $X_{res}$ and $Y_{res}$ respectively, as describe in \eqref{eq:z' calculate}.

\begin{equation}
\mathbf{{z}'} \; \forall \; \mathbf{Pz}_{(i,j)} : \left\lbrace 
    \begin{array}{ll}
        i\:=\left\lfloor \left ( 0,\:\dots\:,\phi'_{max}  \right) \cdot Y_{res} \right \rceil  \\ \\
        j=\left\lfloor \left ( 0,\:\dots\:,\theta'_{max}\right) \cdot X_{res} \right \rceil  \\ 
    \end{array}
    \right.
\label{eq:z' calculate}
\end{equation}

\begin{figure*}[htbp]
    \centering
    \includegraphics[width=\linewidth]{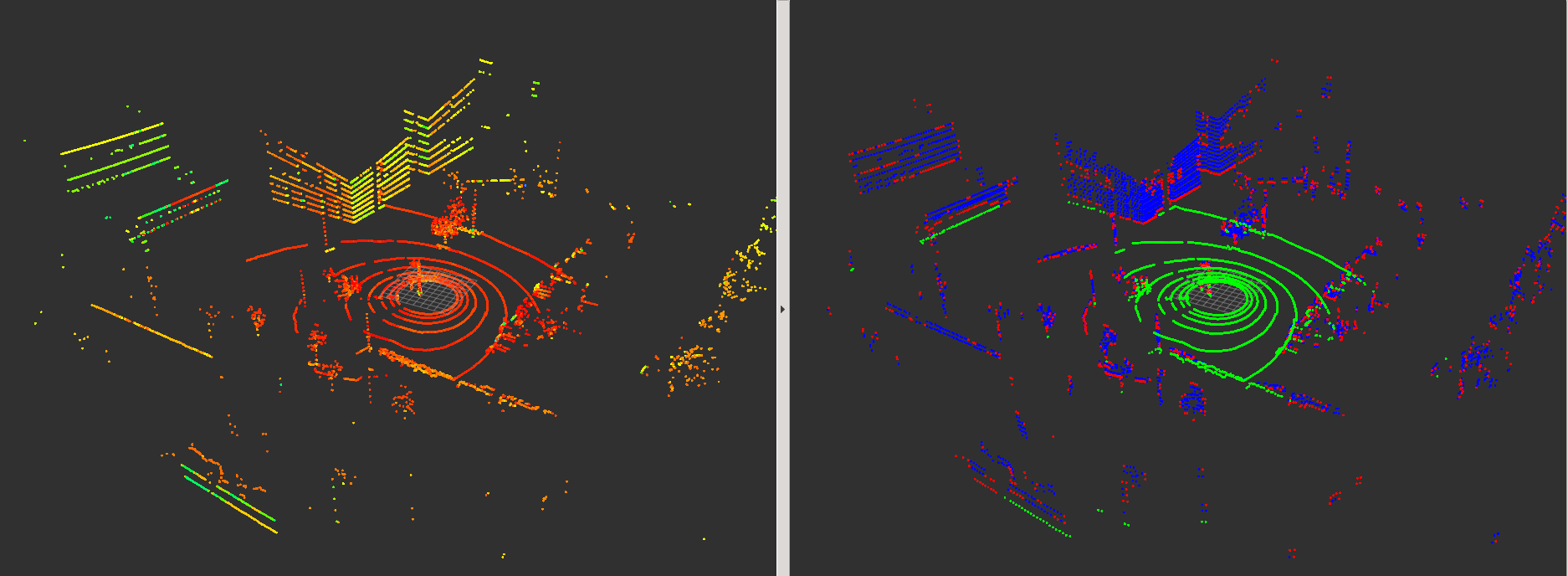}
    \caption{Segmentation of features in a point cloud applying the ground segmentation method of Section \ref{section:Filtering in the frequency domain}. (\textbf{left}) Raw point cloud of Velodyne VLP-16. (\textbf{right}) Point cloud segmentation: ground $\mathbf{P}_{\mathcal{G}}$(green), edge $\mathbf{P}_{\mathcal{E}}$(red) and surface $\mathbf{P}_{\mathcal{S}}$(blue)}
    \label{fig:pointcloudReconstruction}
\end{figure*}

In this way, by obtaining the $\mathbf{x},\mathbf{y},\mathbf{z}$ coordinates of each feature, the $\mathbf{P}_\mathcal{E}$, $\mathbf{P}_\mathcal{S}$ and $\mathbf{P}_\mathcal{G}$ arrays are segmented (\ref{eq:Pe,Ps,Pg arrays}), which correspond to the edge, surface and ground features of the $\mathbf{P}_k$ \eqref{eqn:PointCloud representation} point cloud data. Fig. \ref{fig:pointcloudReconstruction} shows the feature segmentation when using the Fourier mask filtering method.
\begin{equation}
    \begin{matrix}
        \mathbf{P}_{\mathcal{E}} = \{\mathbf{x}_{\mathcal{E}},\mathbf{y}_{\mathcal{E}},\mathbf{z}_{\mathcal{E}}\} \in \mathbb{R}^3\\
        \mathbf{P}_{\mathcal{S}} = \{\mathbf{x}_{\mathcal{S}},\mathbf{y}_{\mathcal{S}},\mathbf{z}_{\mathcal{S}}\} \in \mathbb{R}^3\\
        \mathbf{P}_{\mathcal{G}} = \{\mathbf{x}_{\mathcal{G}},\mathbf{y}_{\mathcal{G}},\mathbf{z}_{\mathcal{G}}\} \in \mathbb{R}^3\\
    \end{matrix}
    \label{eq:Pe,Ps,Pg arrays}
\end{equation}
\section{POSE ESTIMATION}
\label{section:Pose estimation}
The LOAM method aligns the current edge $\mathbf{P}_\mathcal{E}$ and surface $\mathbf{P}_\mathcal{S}$ features with a local feature maps  $\mathbf{P}^l_\mathcal{E}$,  $\mathbf{P}^l_\mathcal{S}$ to pose estimation. For point-to-edge and point-to-plane data matching, edge features are defined as points, and surface and ground features are defined as planes. To generate a local map for data matching, we store the edge and surface features in \mbox{KD-Trees}. Minimizing the distance between characteristic points allows estimating the optimal pose of the current frame and the local map generated in each iteration, as shown in (\ref{eq:minimizer}).
\begin{equation}
    \mathbf{T^*_k} = \operatorname*{argmin}_{\mathbf{T_k}} \sum f_\mathcal{E}(\mathbf{P}_\mathcal{E}) + \sum f_\mathcal{S}(\mathbf{P}_\mathcal{S}) \\
    \label{eq:minimizer}
\end{equation}
Where, $f_\mathcal{E}(\mathbf{P}_\mathcal{E})$ are the distances from the edge features $\mathbf{P}_\mathcal{E}$ to their local map $\mathbf{P}^l_\mathcal{E}$, defined in (\ref{eq:feature edge and local map distances}).
\begin{equation}
    f_\mathcal{E}(\mathbf{P}_\mathcal{E}) = \mathbf{P}_n \cdot ((\mathbf{T}_k\:\mathbf{P}_\mathcal{E}-\mathbf{P}^l_\mathcal{E})\times \mathbf{n}^l_\mathcal{E})\\
    \label{eq:feature edge and local map distances}
\end{equation}
$\mathbf{P}_n$ (\ref{eq:Pn unit vector}) is the unit vector of the function, and $\mathbf{n}^l_\mathcal{E}$ are the eigenvalues associated with the largest value of the orientation and line position of the local edge map. 
\begin{equation}
    \mathbf{P}_n = \frac{(\mathbf{T}_k\:\mathbf{P}_\mathcal{E}-\mathbf{P}^l_\mathcal{E})\times \mathbf{n}^l_\mathcal{E})}{||(\mathbf{T}_k\:\mathbf{P}_\mathcal{E}-\mathbf{P}^l_\mathcal{E})\times \mathbf{n}^l_\mathcal{E}||}\\
    \label{eq:Pn unit vector}
\end{equation}
The norm of the local surface is taken as the eigenvector associated with the smallest eigenvalue $\mathbf{n}^l_\mathcal{S}$. The distance between the surface features and local map surface, $f_\mathcal{S}(\mathbf{P}_\mathcal{S})$, is defined in (\ref{eq:feature surf and local map distances}).
\begin{equation}
    f_\mathcal{S}(\mathbf{P}_\mathcal{S}) = (\mathbf{T}_k\:(\mathbf{P}_\mathcal{S}+\mathbf{P}_\mathcal{G})-\mathbf{P}^l_\mathcal{S})\cdot \mathbf{n}^l_\mathcal{S})
    \label{eq:feature surf and local map distances}
\end{equation}

Similar to F-LOAM and \cite{zhang2017robust}, the eigenvectors are calculated with the covariance matrix of their nearby points with their respective local map. In the experiment, the covariance matrix is calculated considering a neighborhood radius of 1.0 meter. In contrast to F-LOAM, where the ground and surface features are analyzed together, we have divided the surface feature in (\ref{eq:feature surf and local map distances}) into  three point cloud groups ($\mathbf{P}_\mathcal{S}$, $\mathbf{P}_\mathcal{G}$ and $\mathbf{P}_\mathcal{S}\!+\!\mathbf{P}_\mathcal{G}$), and we have evaluated which group of features is suitable for KITTI and for a Velodyne VLP-16 sensor. The pose estimation can be derived by solving the non-linear equation through Gauss-Newton method. The current position $\mathbf{T}^*_k$ (\ref{eq:minimizer}) is obtained by solving the nonlinear optimization. To correct for the existing distortion when estimating position with point clouds, we use the method shown in \cite{wang2021floam}, which first assumes a constant linear and angular velocity over a short period and then the distortion is calculated after the position estimation process. This method achieves lower computational cost than the classical LOAM, so that the position can be estimated at high frequencies. In addition, by recombining the point cloud and segmenting it by features (see Section \ref{pcl_reconstruction}), each local map can be reduced using a 3D voxelized grid implemented with the point cloud library PCL \cite{rusu20113d}. In this way, the number of segmented features is reduced and the pose estimation has a lower computational cost.

\section{Experiments}
In this section we present experiments and results of our \mbox{LiLO} odometry method. This method has been implemented on an intel 2.60GHz 6-core processor (the method uses only 1-core ) and the testing environment has been based on ROS Melodic and Ubuntu 18.04. Experiments on different point cloud groups are analyzed separately: \textit{EG}, (edges and ground), \textit{ES} (edges and surfaces) and \textit{EGS} (edges, ground and surfaces). Moreover, we have analyzed the results obtained from different SRI sizes. The performance has been evaluated by means of the execution time and the number of features extracted for pose estimation, which is directly proportional to the computational cost. Also, we verified the results of translation and rotation error rates using the evaluation tool presented in \cite{kitti_eval}, where the average translation error (ATE) and average rotation error (ARE) are calculated and defined in \cite{KITTI_Benchmark}. The point clouds of the local maps $\mathbf{P}^l_\mathcal{E}$ and $\mathbf{P}^l_\mathcal{S}$ were continuously cleaned by passing a radius of 100 m in the $x, y, z$ axes.

\begin{figure}[!b]
    \centering
     \includegraphics[clip, width=\linewidth]{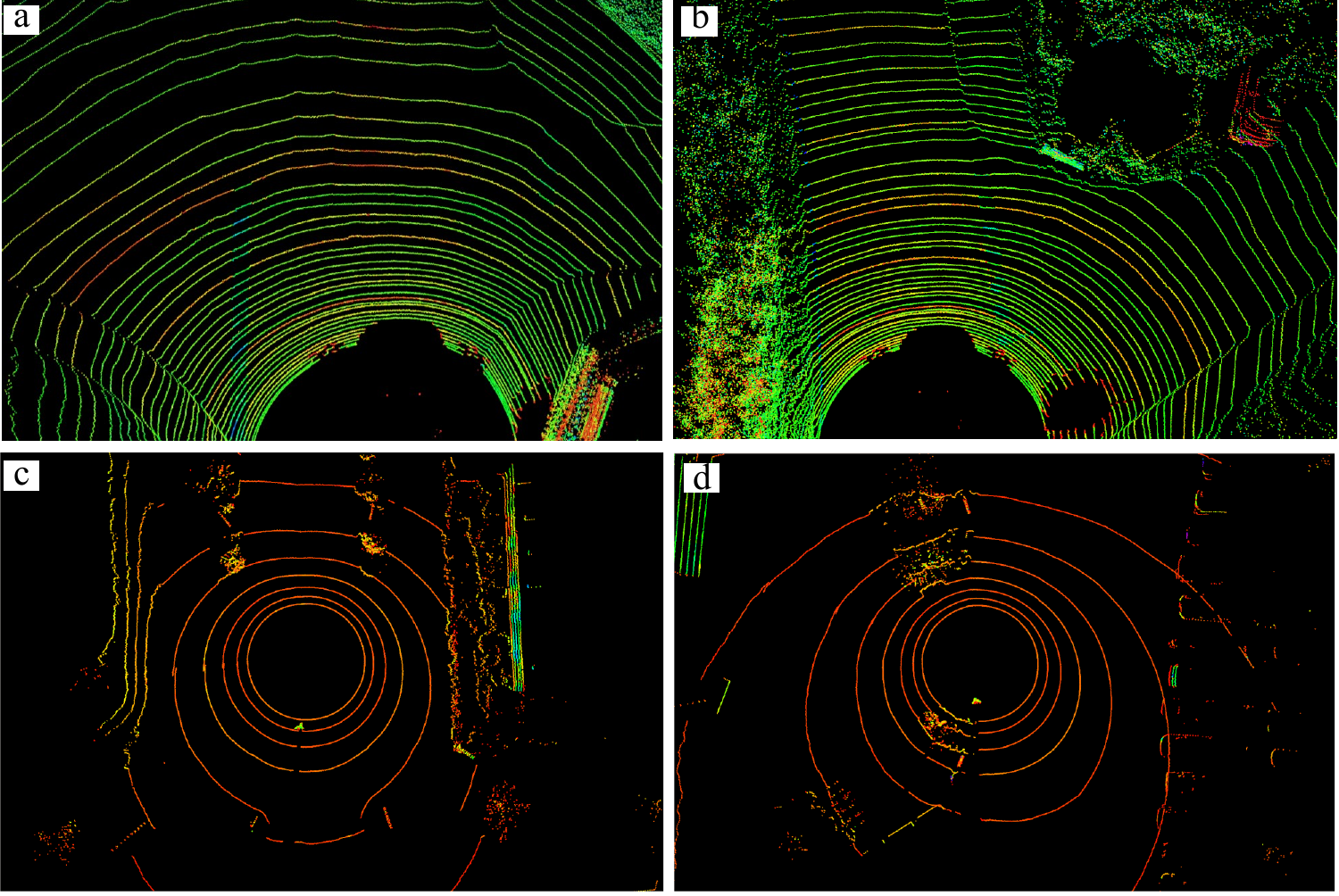}
    \caption{Bird's eye view of the point cloud of the KITTI dataset and the Velodyne VLP-16 sensor. \textbf{(a)} Sequence 05 KITTI. \textbf{(b)} Sequence 06 KITTI. \textbf{(c)} Loop 1. \textbf{(d)} Loop 3.}
    \label{fig:ground_features}
\end{figure}

We have evaluated which method of extracting ground features is the most suitable for the KITTI dataset sensors and for a Velodyne VLP-16 sensor. Fig. \ref{fig:ground_features}a and Fig. \ref{fig:ground_features}b shows a fragment of the point cloud of sequences 05 and 09 of the KITTI dataset, where each point of the LiDAR sensor light beams projected on the ground has different distances. This does not generate a circular shape when transforming the point cloud into a SRI, instead, it generates a shade that, when passed to the frequency domain, is composed of several frequencies. Thus, it is not possible to segment the ground plane of the KITTI dataset point cloud with the frequency domain filtering method. That is why, for the KITTI dataset, we use the ground feature filtering method of  Section \ref{section:Edge and surface segmentation}, since the Sobel operator calculates an approximation of the gradient of an image intensity function, filtering out the horizontal lines of the ground with noisy range values. For the case of the Velodyne VLP-16 sensor, as shown in Fig. \ref{fig:ground_features}c and Fig. \ref{fig:ground_features}d, the point cloud projected on the ground does not generate noise and these points are projected in a circular shape in the SRI, making possible the frequency domain filtering shown in Section \ref{section:Filtering in the frequency domain}.

\begin{figure}[htbp]
    \centering
     \includegraphics[clip, width=\linewidth]{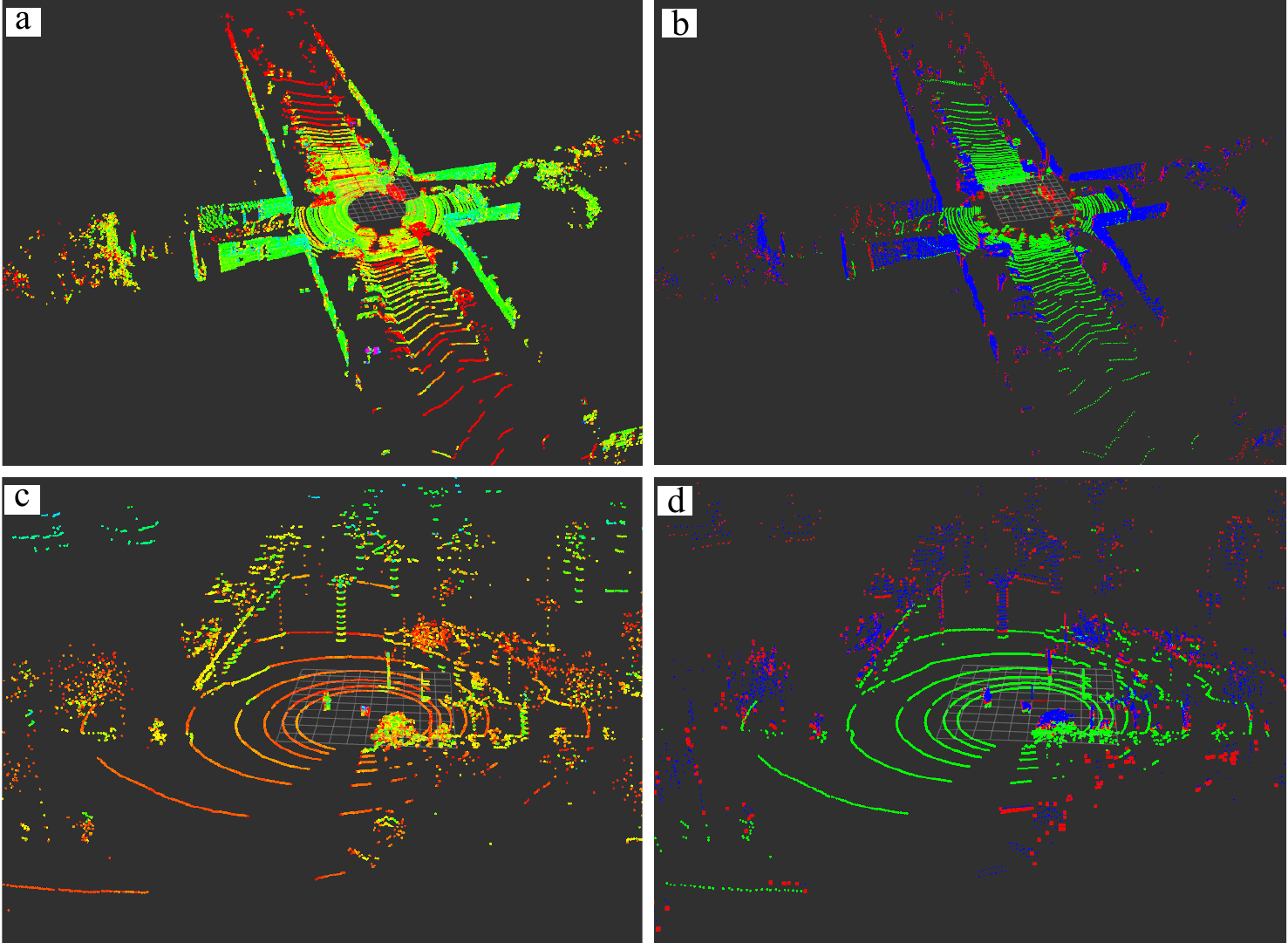}
    \caption{Segmentation of edge, surface and ground features from a sequence of the KITTI dataset \textbf{(a)(b)} and a point cloud given by the Velodyne VLP-16 \textbf{(c)(d)}. }
    \label{fig:segmentation_features}
\end{figure}

Fig. \ref{fig:segmentation_features}a and Fig. \ref{fig:segmentation_features}b shows the feature segmentation results of Edges, Surfaces and Ground point cloud from the KITTI dataset with Sobel operator. Fig. \ref{fig:segmentation_features}c and Fig. \ref{fig:segmentation_features}d, on the other hand, shows the point cloud feature segmentation of Velodyne VLP-16 sensor from BLUE robotic platform. 

\subsection{KITTI dataset evaluation}
\label{sec:KITTI dataset evaluation}

\begin{figure*}[htbp]
 \centering
  \subfloat[Sequence 01\label{fig:kitti_features_01}]{%
        \includegraphics[width=\linewidth]{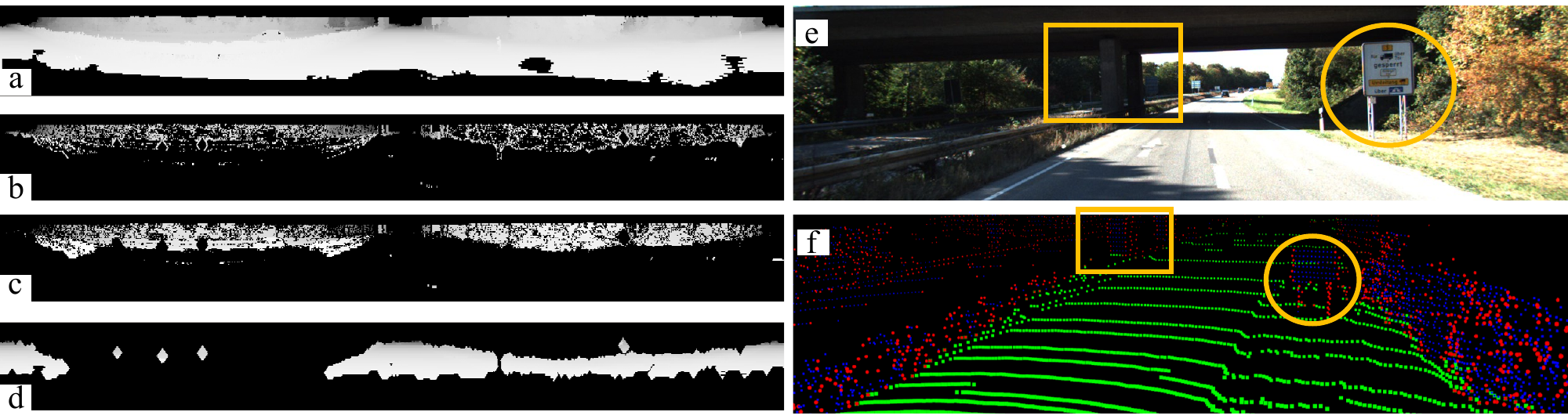}}
  \\
  \subfloat[Sequence 06\label{fig:kitti_features_06}]{%
       \includegraphics[width=\linewidth]{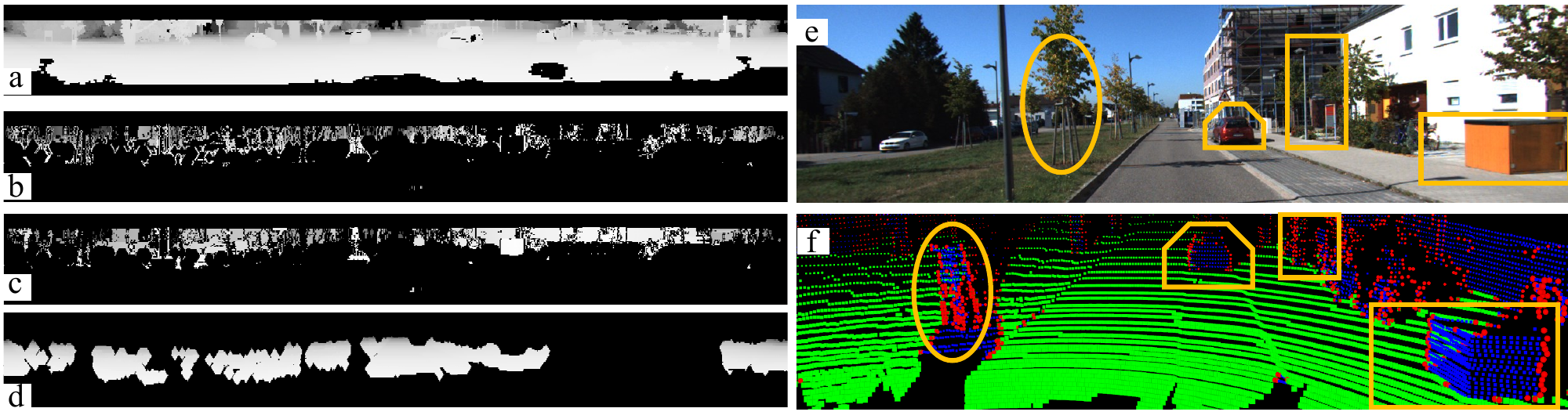}}
 \caption{Feature filtering of the point cloud by converting it into a SRI of the KITTI dataset. \textbf{(a)} SRI. \textbf{(b)} Edge features. \textbf{(c)} Surface features. \textbf{(d)} Ground features. \textbf{(e)} RGB image corresponding to each sequence of the KITTI dataset. \textbf{(f)} Segmented point cloud of the corresponding sequence of the KITTI dataset.} 
\end{figure*}

First, we evaluated the robustness of feature segmentation on the KITTI dataset with the Sobel operator method of Section \ref{section:Edge and surface segmentation}. Fig. \ref{fig:kitti_features_01} and Fig. \ref{fig:kitti_features_06} show the feature segmentation of sequences 01 and 06 of the KITTI dataset compared to the RGB images of their corresponding scenario. Then, we evaluated the performance and runtime on 11 sequences (00-10) of the KITTI dataset described in \cite{KITTI_Benchmark}, where the 3D point scans were collected from the Velodyne HDL-64E and recorded at 10 Hz. The 11 sequences contained data from urban, rural, and highway environments. The KITTI sequence experiments were done with different groups of point clouds and image sizes. When converting a LiDAR point cloud into a SRI, the images are created with an height equal to the number of LiDAR channels, and a width corresponding to the 360º range in factors: 1.0, 0.5 and 0.35. This parameters given images of 64x360, 64x720 and 64x1024 pixels respectively. 

\begin{figure*}[htbp]
    \centering

    \subfloat[Groups of point clouds by features \label{fig:EG_EGS_ES_different_Size}]{
    \includegraphics[clip, trim=0cm 4cm 0cm 4cm, width=0.4\linewidth]{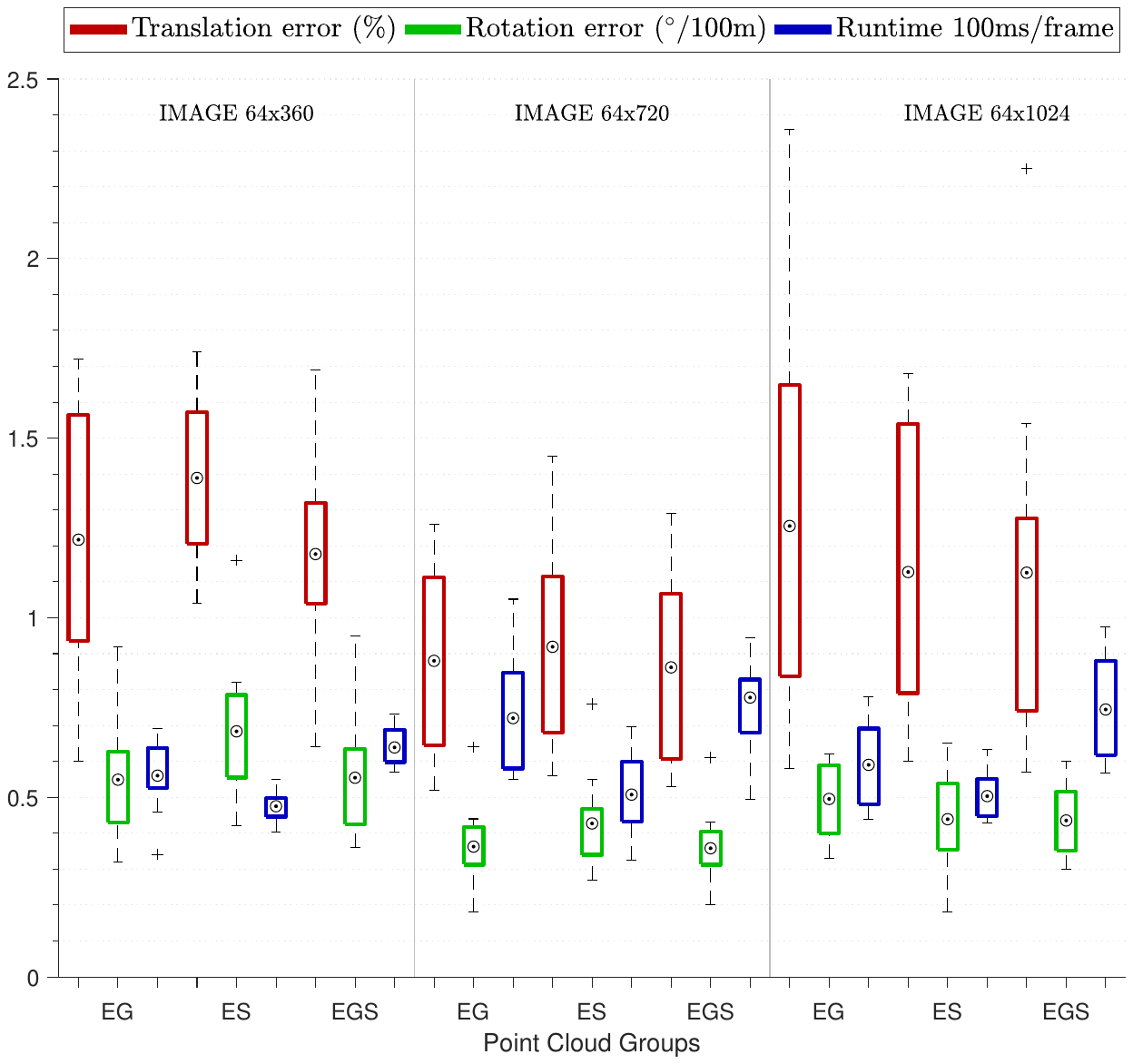}}
    \subfloat[Pose estimation results for Sequence 01\label{fig:fig_kitti_01}]{
    \includegraphics[width=0.6\linewidth]{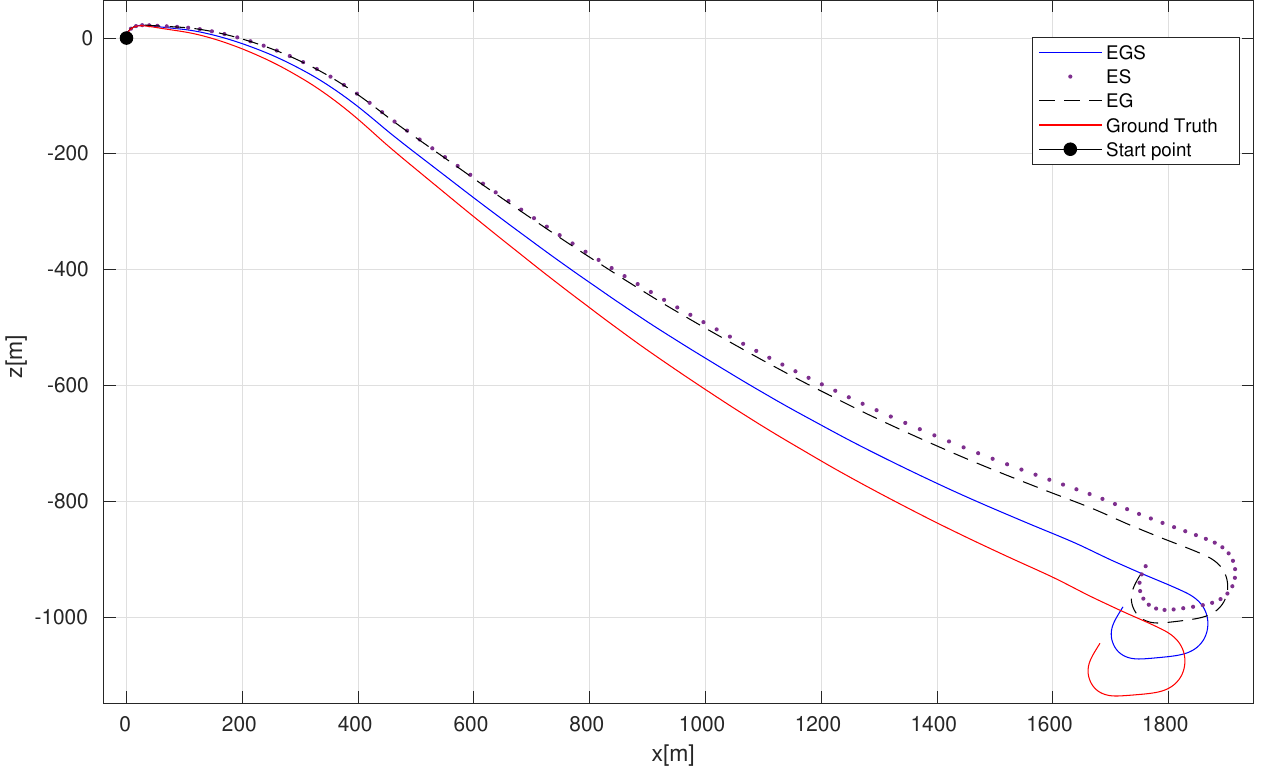}}

\caption{\textbf{(a)} Experiments with different image size and Point Cloud groups. Red boxes are the translation error, green boxes are the rotation error, and blue boxes are the runtime scaled to 100ms. Average of each boxplot is represent by $\odot$, while cross symbols $+$ represent the outliers. \textbf{(b)} Pose estimation results for Sequence 01 of the KITTI dataset with different point cloud groups.}
\end{figure*}

The SRI is filtered using the Sobel algorithm, as detailed in Section \ref{section:Edge and surface segmentation}. Besides, the experiments have different groups of point clouds, where the edge features $\mathbf{P}_\mathcal{E}$ are matched with the edge map $\mathbf{P}^l_\mathcal{E}$ and the surface features are divided into: only surface points $\mathbf{P}_\mathcal{S}$, only ground points $\mathbf{P}_\mathcal{G}$, or these together $\mathbf{P}_\mathcal{S}\!+\!\mathbf{P}_\mathcal{G}$. Then, surface features are matched with the surface map $\mathbf{P}^l_\mathcal{S}$. Experiments with only edge point clouds are not analyzed, as detailed in \cite{Dilo}, the overall estimation performance is degraded using only edge point clouds. In this way, we analyzed the rotation and translation errors with different point cloud groups in the 11 KITTI sequences as shown in Fig. \ref{fig:EG_EGS_ES_different_Size}, where the average values of ATE, ARE and runtime are shown as $\mathbf{\odot}$ in each \textit{boxplot}, and the cross symbols represent the outliers of the experiments.  Most of these outliers are produced by sequence 01, which could be because this experiment present a wide path with very few structures in a road environment.

With an image of 64x720 pixels, the translation and rotation percentages are lower compared to 64X360 and 64x1024 images. In the experiments, the processing time are within the range of 10Hz which is the execution time limit.

\begin{figure*}[htbp]
\centering

\subfloat[00 \label{fig:fig_kitti_00}]{%
        \includegraphics[width=0.195\textwidth]{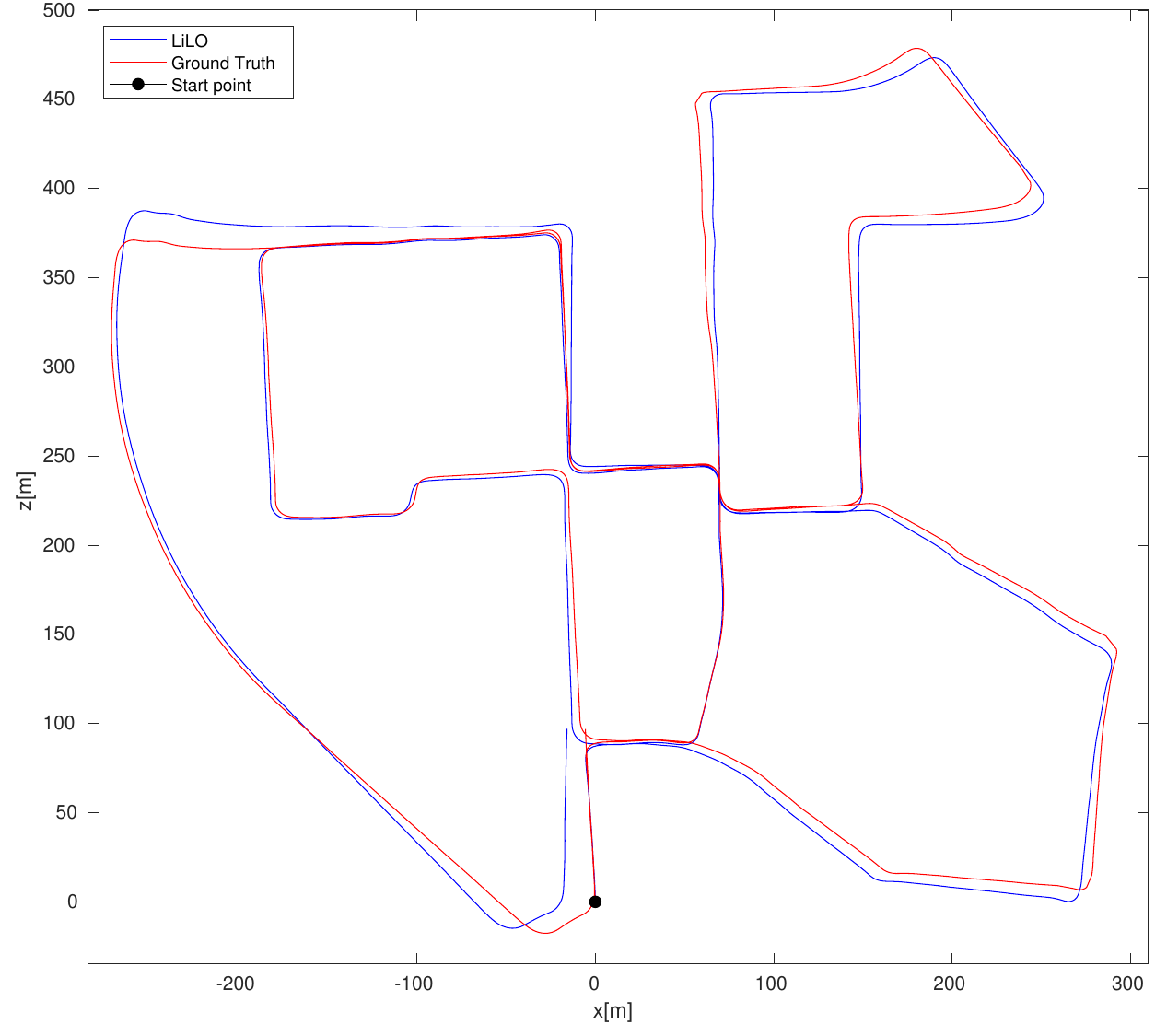}}
\hfill
\subfloat[02 \label{fig:fig_kitti_02}]{%
        \includegraphics[width=0.195\textwidth]{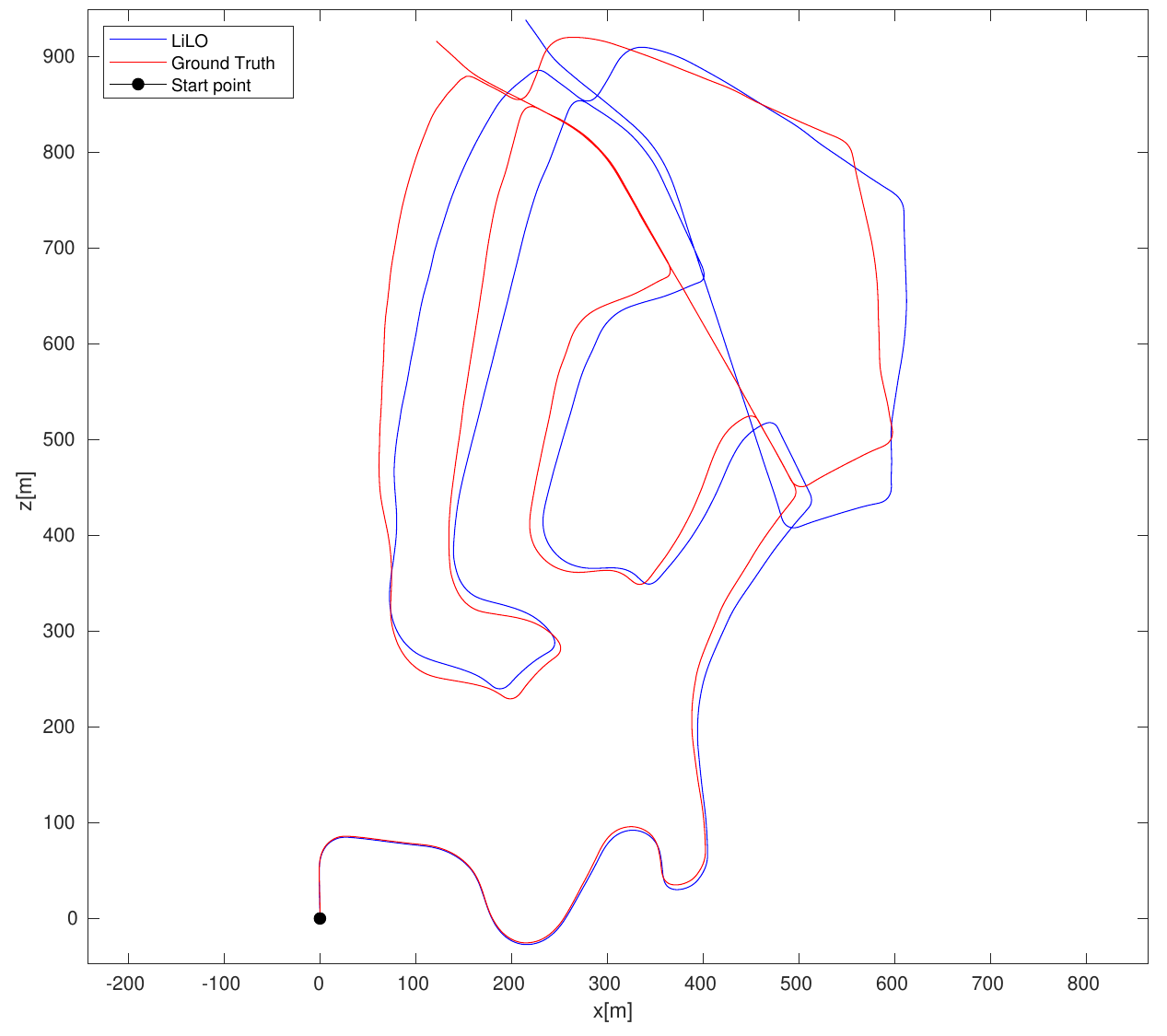}}
\hfill
\subfloat[03 \label{fig:fig_kitti_03}]{%
        \includegraphics[width=0.195\textwidth]{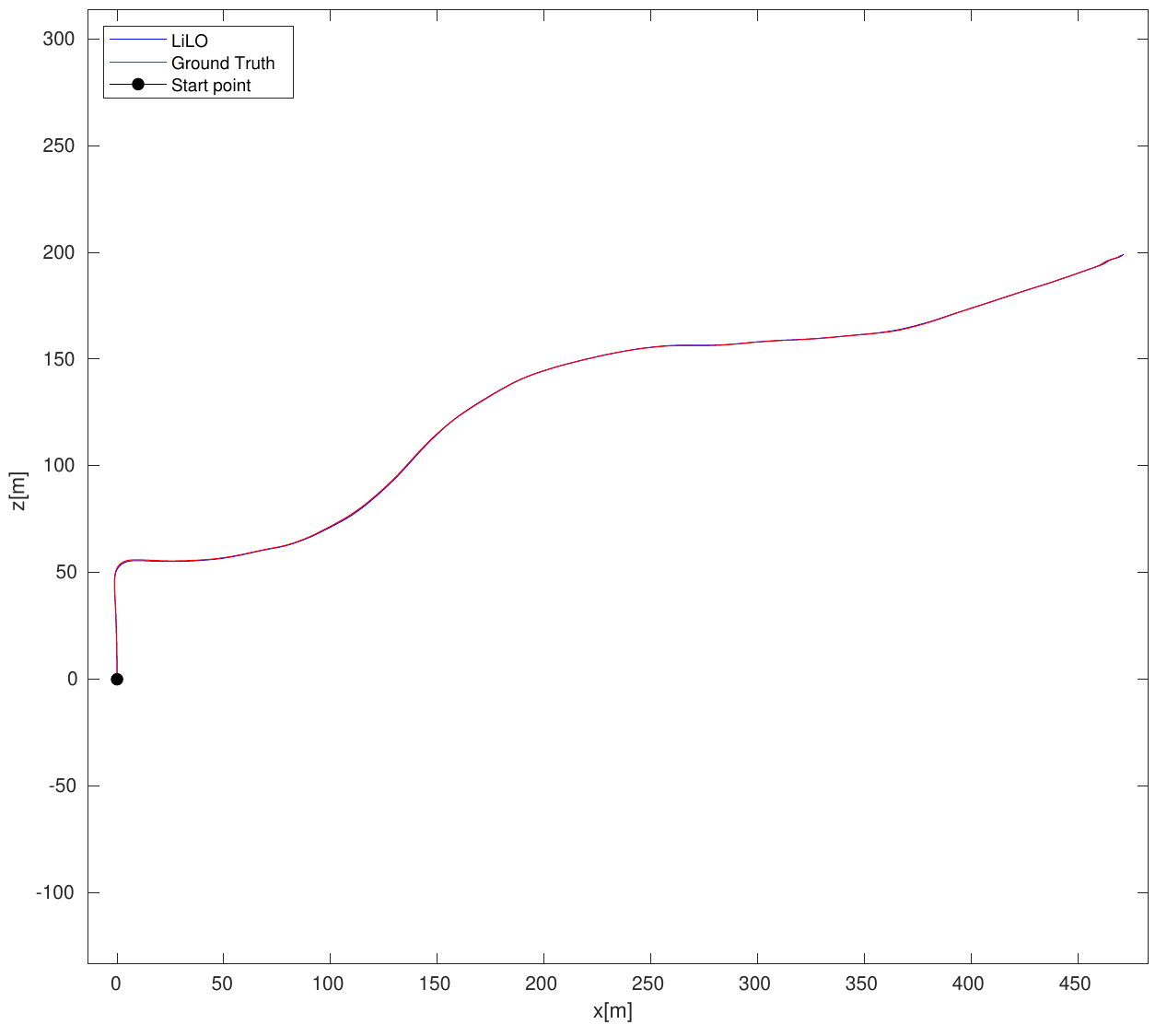}}
\hfill
\subfloat[04 \label{fig:fig_kitti_04}]{%
        \includegraphics[width=0.195\textwidth]{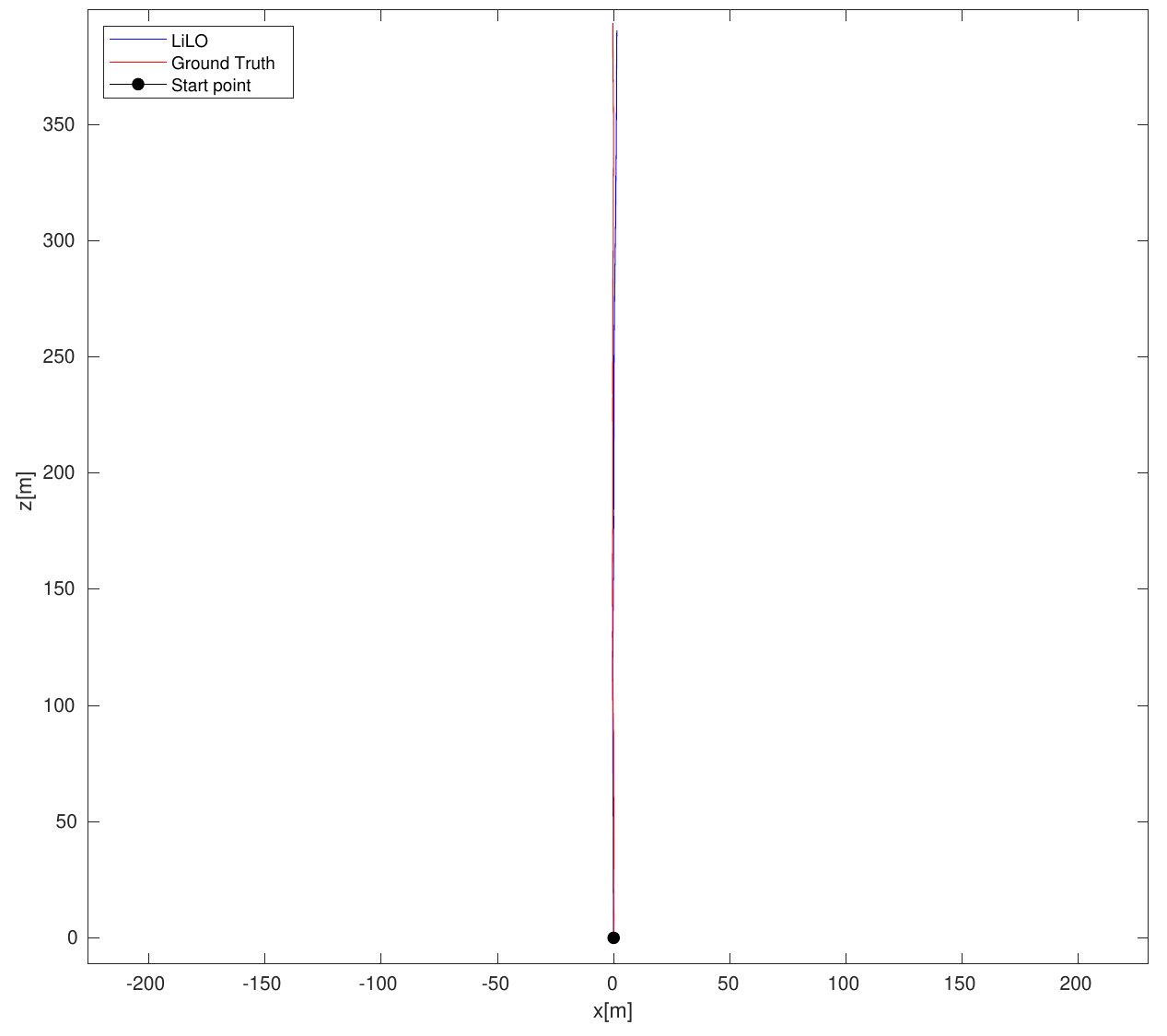}}
\hfill
\subfloat[05 \label{fig:fig_kitti_05}]{%
        \includegraphics[width=0.195\textwidth]{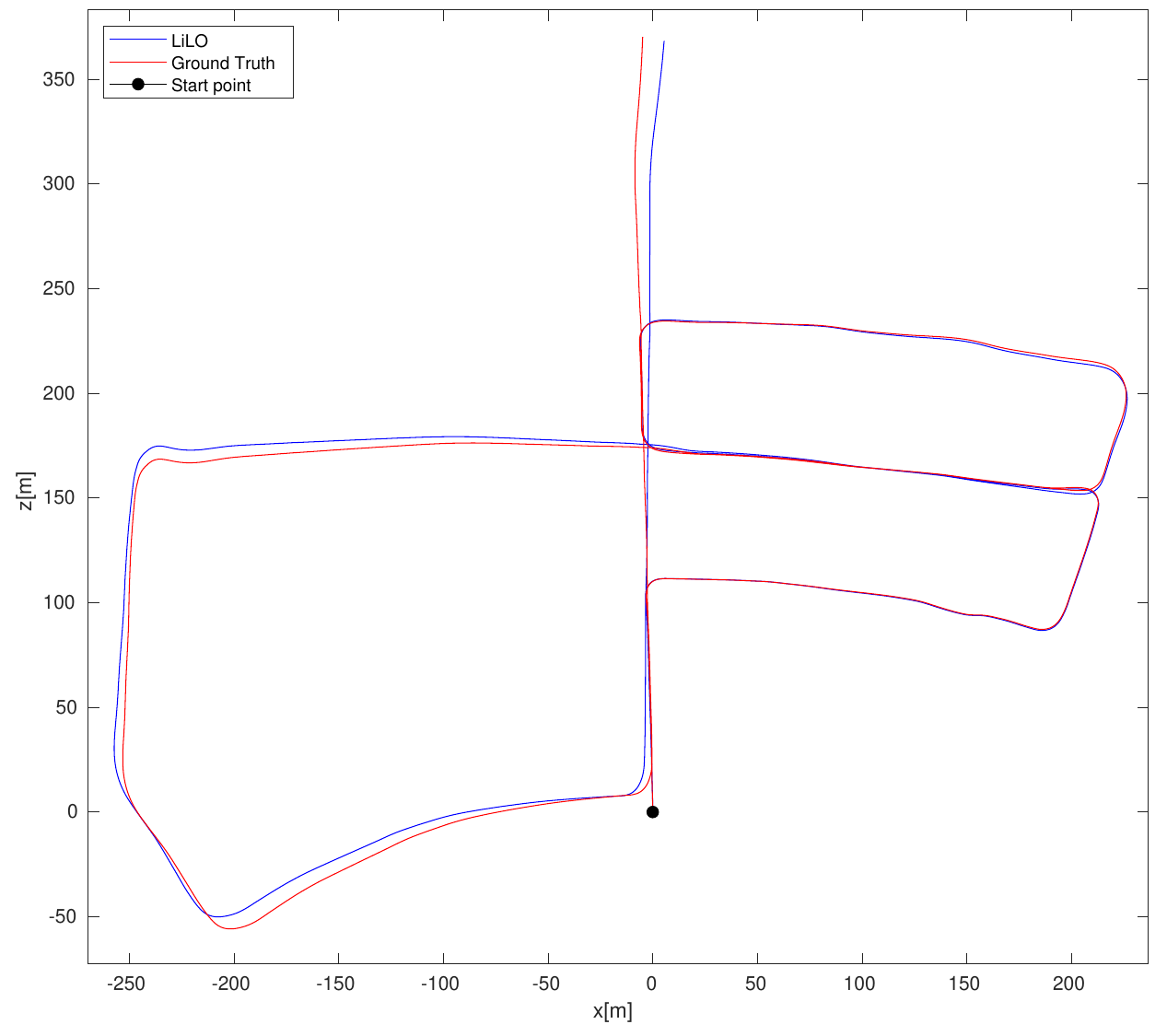}}
\\

\subfloat[06 \label{fig:fig_kitti_06}]{%
        \includegraphics[width=0.195\textwidth]{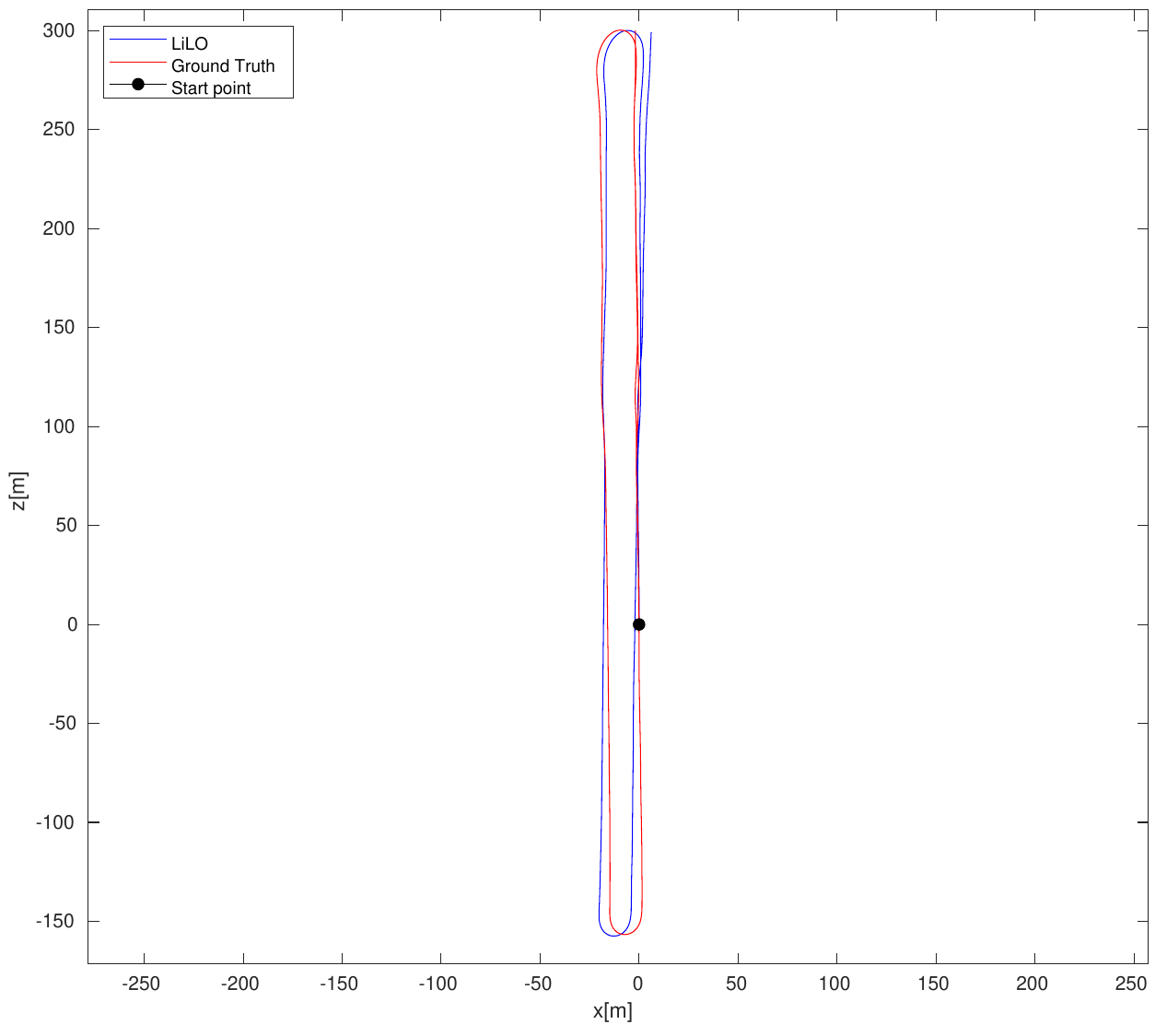}}
\hfill
\subfloat[07 \label{fig:fig_kitti_07}]{%
        \includegraphics[width=0.195\textwidth]{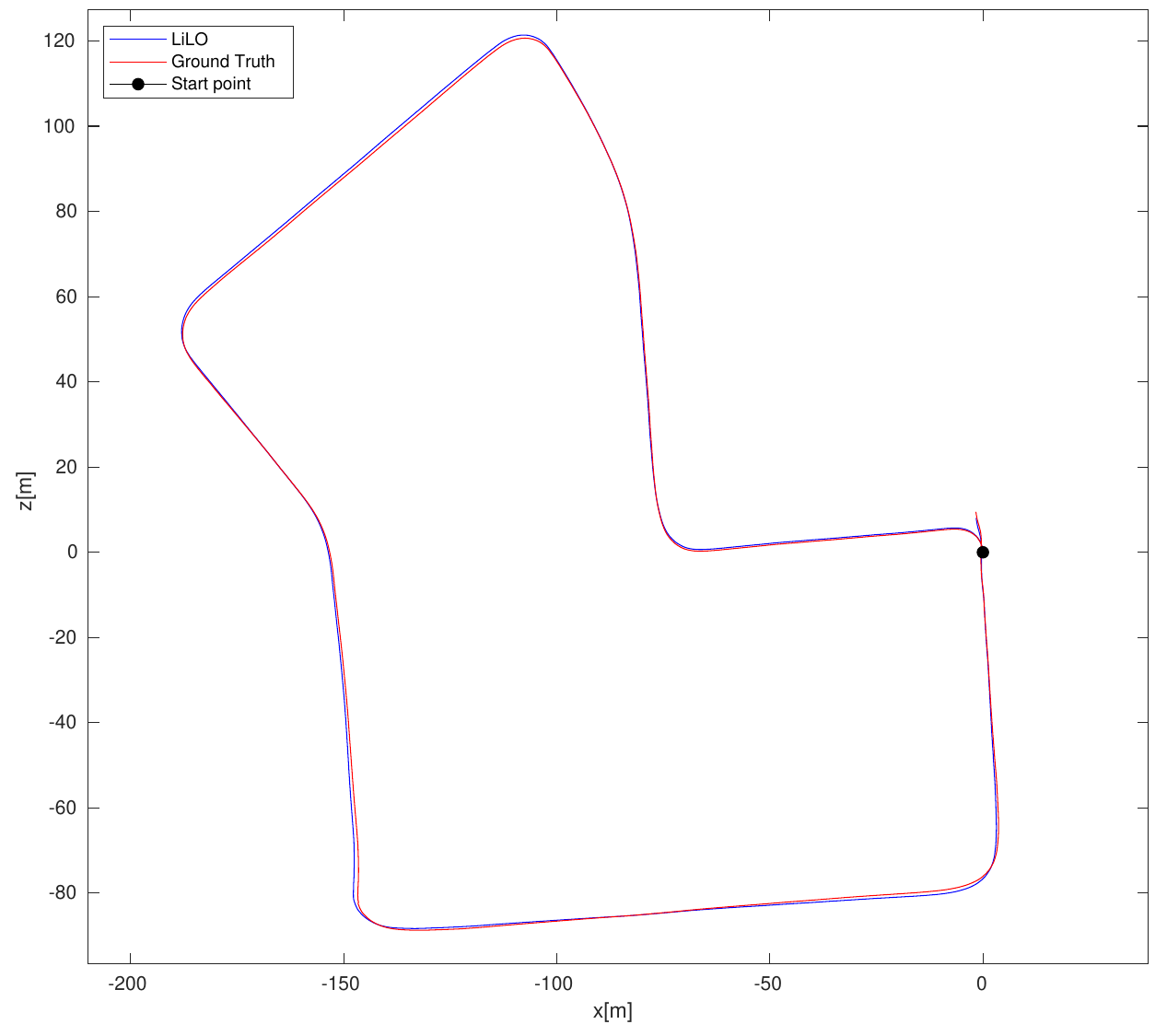}}
\hfill
\subfloat[08 \label{fig:fig_kitti_08}]{%
        \includegraphics[width=0.195\textwidth]{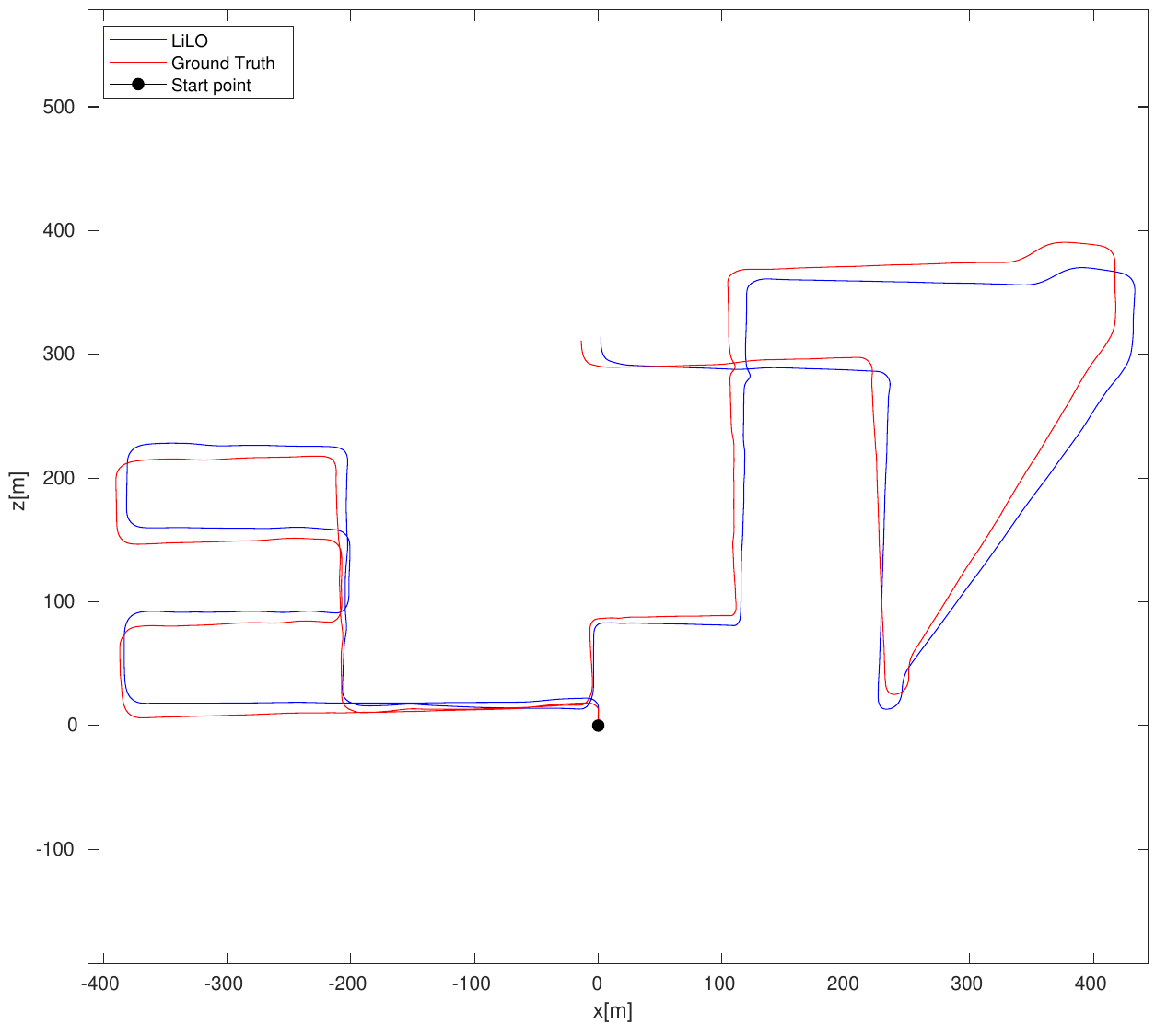}}
\hfill
\subfloat[09 \label{fig:fig_kitti_09}]{%
        \includegraphics[width=0.195\textwidth]{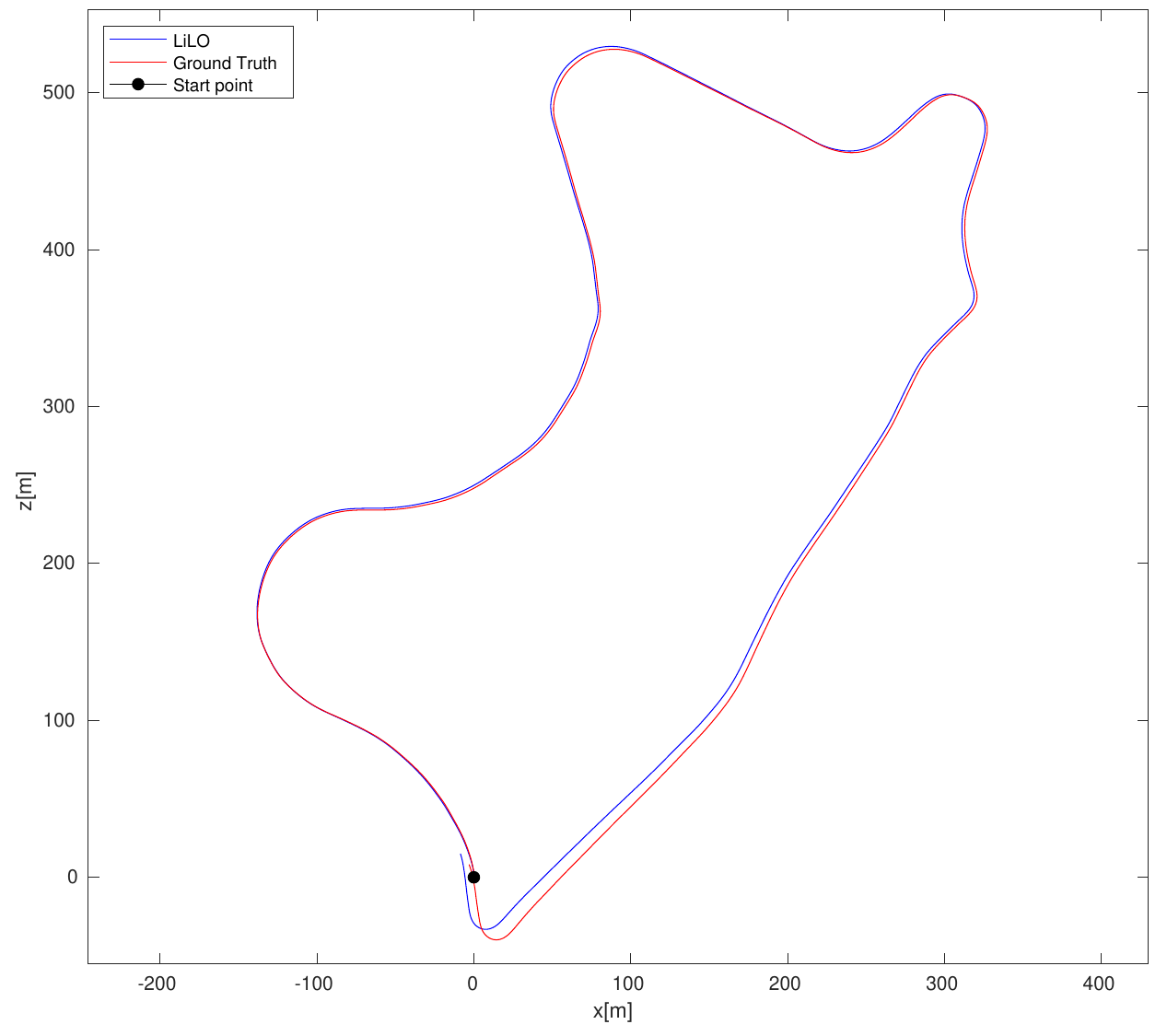}}
\hfill
\subfloat[10 \label{fig:fig_kitti_10}]{%
        \includegraphics[width=0.195\textwidth]{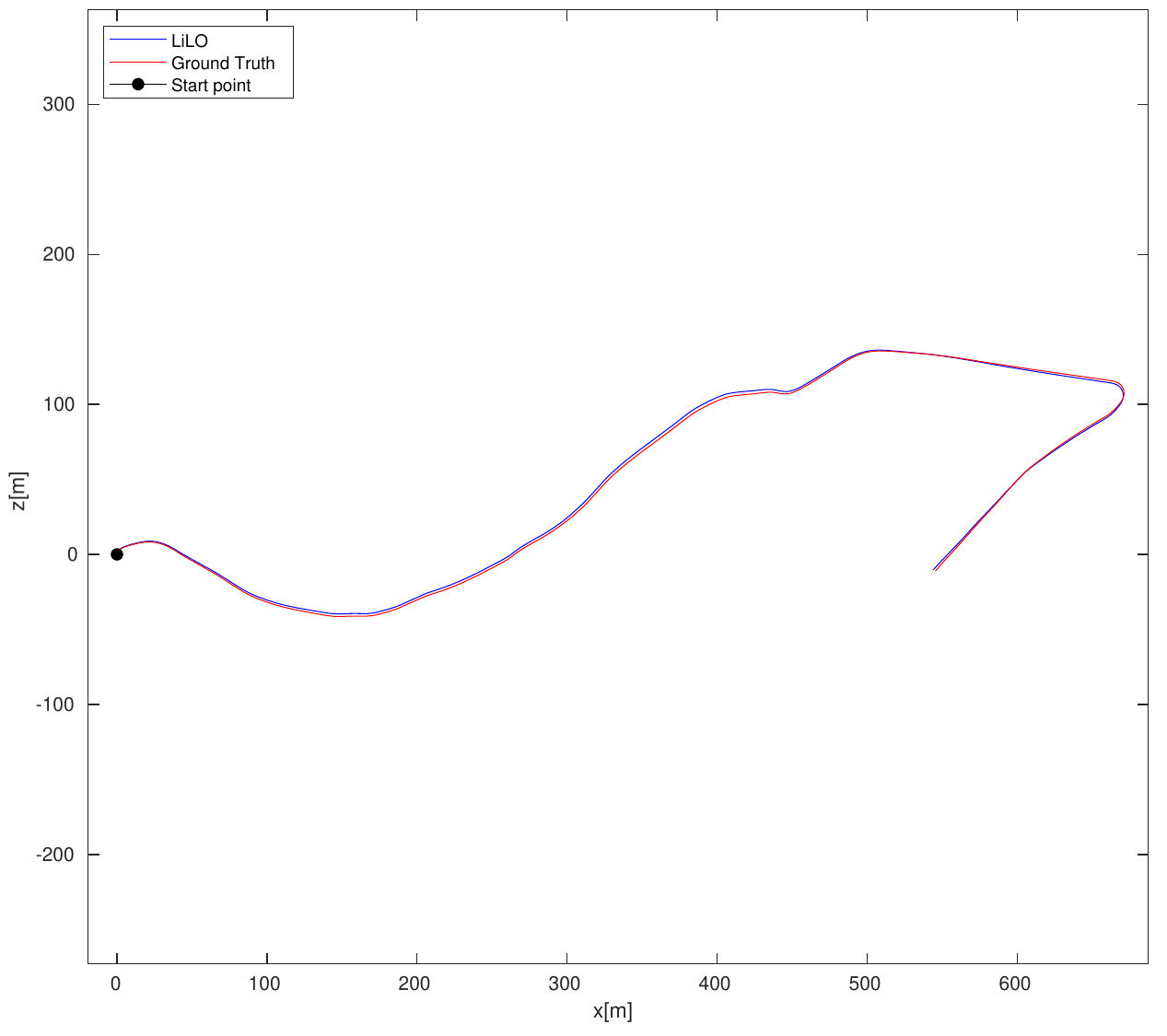}}

\caption{Result of the proposed method on the sequences of KITTI dataset (00,02-10). The blue line is the estimated odometry and the red line is the ground truth based on GPS data.}
\label{fig:fig_kitti_00-10}
\end{figure*} 

\begin{table*}
   \footnotesize
   \centering
    \caption{Accuracy analysis for the KITTI odometry dataset. Each result is represented as translation(\%)/rotation(º/100m). The values of each method are taken without loop closure algorithms. The average translation error (ATE) and average rotation error (ARE) are calculated and defined in \cite{KITTI_Benchmark}. LOAM values were taken from \cite{lodonet}, T-LOAM and F-LOAM from \cite{T-LOAM}, and LEGO-LOAM and LiODOM results from \cite{garcia2022liodom}. Best results are show in bold red and second best in blue.}
    \begin{tabular}{lllllllll}
    \toprule
    \begin{tabular}[c]{@{}c@{}}\textbf{Seq.}\\\textbf{No}\end{tabular} & \begin{tabular}[c]{@{}c@{}}\textbf{Path}\\\textbf{Len.}\textbf{(m)}\end{tabular} & \textbf{Environment} & \textbf{LOAM}& \textbf{T-LOAM} & \begin{tabular}[c]{@{}c@{}}\textbf{LeGO-}\\\textbf{LOAM}\end{tabular} & \textbf{F-LOAM} & \textbf{LiODOM} & \textbf{LiLO}  \\ 
    \hline
    \textbf{00}& 3714& Urban                              & \textcolor{blue}{0.78}/0.53              & 0.98/0.60              & 2.17/1.05              & 1.11/\textcolor{blue}{0.40}              & 0.86/\textcolor{red}{0.35}              & \textcolor{red}{0.71}/0.43       \\
    \textbf{01}& 4268& Highway                            & 1.43/0.55              & 2.09/0.52              & 13.4/1.02              & 3.01/0.85              & \textcolor{blue}{1.30}/\textcolor{red}{0.13}              & \textcolor{red}{1.29}/\textcolor{blue}{0.20}       \\
    \textbf{02}& 5075& Urban+Country                      & \textcolor{red}{0.92}/0.55              & 1.01/0.39              & 2.17/1.01              & 1.22/0.43              & \textcolor{blue}{0.95}/\textcolor{red}{0.31}              & 1.06/\textcolor{blue}{0.39}       \\
    \textbf{03}& 563& Country                             & \textcolor{red}{0.86}/0.65              & \textcolor{blue}{1.10}/0.82              & 2.34/1.18              & 4.51/1.84              & 1.26/\textcolor{red}{0.23}              & 1.22/\textcolor{blue}{0.31}       \\
    \textbf{04}& 397& Country                             & \textcolor{blue}{0.71}/0.50              & \textcolor{red}{0.68}/0.68              & 1.27/1.01              & 0.93/0.63              & 1.41/\textcolor{red}{0.01}              & 0.84/\textcolor{blue}{0.28}       \\
    \textbf{05}& 2223& Country                            & 0.57/0.38              & \textcolor{blue}{0.55}/\textcolor{red}{0.32}             & 1.28/0.74              & 0.63/\textcolor{blue}{0.32}              & 0.83/0.36              & \textcolor{red}{0.53}/0.34       \\
    \textbf{06}& 1239& Urban                              & 0.65/0.39              & \textcolor{blue}{0.56}/\textcolor{red}{0.31}              & 1.06/0.63              & 2.15/0.74              & 0.83/0.33              & \textcolor{red}{0.54}/\textcolor{blue}{0.32}       \\
    \textbf{07}& 695& Urban                               & 0.63/0.50              & \textcolor{red}{0.50}/\textcolor{blue}{0.47}              & 1.12/0.81              & \textcolor{blue}{0.51}/\textcolor{red}{0.35}              & 0.88/0.61              & 0.60/0.61       \\
    \textbf{08}& 3225& Urban+Country                      & 1.12/0.44              & \textcolor{blue}{0.94}/\textcolor{blue}{0.33}              & 1.99/0.94              & 0.97/0.37              & \textcolor{red}{0.86}/\textcolor{red}{0.33}              & 1.07/0.41       \\
    \textbf{09}& 1717& Urban+Country                      &\textcolor{blue}{0.77}/0.48              & 0.80/0.40              & 1.97/0.98              & 0.82/0.40              & 1.03/\textcolor{blue}{0.32}              & \textcolor{red}{0.63}/\textcolor{red}{0.32}        \\
    \textbf{10}& 919& Urban+Country                       & \textcolor{red}{0.79}/0.57              & 1.12/0.61              & 2.21/0.92              & 2.52/0.96              & 1.20/\textcolor{red}{0.29}              & \textcolor{blue}{0.99}/\textcolor{blue}{0.33}       \\
    \textbf{avg}& -    & -                                & \textcolor{red}{0.84}/0.50              & 0.93/0.49              & 2.82/0.94              & 1.67/0.66              & 1.04/\textcolor{red}{0.30}              & \textcolor{blue}{0.86}/\textcolor{blue}{0.36}            \\
    \hline
    \end{tabular}
    \label{tab:Kitti table odometry comparation}
\end{table*}

\begin{table}[tbp]
\footnotesize
\centering
\caption{Runtime and environments of each Lidar odometry method with which our method is compared. The LeGo-LOAM show in the table. The LeGO-LOAM* method shown in the table was taken from \cite{KITTI_Benchmark} and is the frame-by-frame version of pose estimation. The LeGO-LOAM* error is 19.57\% in ATE and 4.93°/100 m in ARE evaluated on the KITTI dataset. All methods are developed in C/C++.}

\begin{tabular}{lll}
\textbf{Method} & \textbf{Runtime(ms)} & \textbf{Environment}      \\
\hline
\scriptsize

LOAM            & 100                   & 2 core  \\
T-LOAM          & 82                    & 6 core  \\
LeGO-LOAM*      & 30                    & 1 core  \\
F-LOAM          & 100                   & 1 core  \\
LiODOM          & 59                    & 6 core  \\
LiLO            & 78                    & 1 core  \\

\hline
\end{tabular}
\label{tab:time_and_enviroments}
\end{table}

The deviations in the execution time are caused by the path analyzed and the amount of point clouds per frame. In images of 64x360 pixels, execution time are low due to the smaller number of elements, with the disadvantage of having a higher ATE. The trajectories represented in Fig. \ref{fig:fig_kitti_01} indicate the results of sequence 01 using different surface features shown in (\ref{eq:feature surf and local map distances}) and explained in Section \ref{section:Pose estimation}.  The trajectory generated by an \textit{EGS} point cloud (blue line) is closer to the ground truth dataset (red line), resulting in an ATE of 1.29\% and  an ARE of 0.2°/100 m. Fig. \ref{fig:fig_kitti_00-10} shows the results of our method for each sequence of the KITTI dataset using an \textit{EGS} point cloud group. 

The translation and rotation error are shown in Table \ref{tab:Kitti table odometry comparation}. The LOAM results were taken from \cite{lodonet}, T-LOAM and F-LOAM from \cite{T-LOAM}, and LEGO-LOAM and LIODOM results from \cite{garcia2022liodom}. Compared to LOAM, T-LOAM, LeGO-LOAM, F-LOAM and LIODOM methods without loop closure, our LiLO odometry method has some of the lowest measured errors (0.86\% ATE and 0.36º/100m ARE), only minimally improved by the LOAM method (0.84\% and 0 .5º/100m), but with the difference that LOAM has a 100 ms runtime. Our method has a mean run-time value of 78 ms for the KITTY dataset, which makes it an ideal odometry estimation method for a LiDAR sensor with a frequency of 10 Hz.

Table \ref{tab:time_and_enviroments} shows the runtime and environments for each odometry system with which we compared our method. The data for the LOAM, LeGo-LOAM and F-LOAM odometry systems were taken from the KITTI dataset \cite{KITTI_Benchmark}. For the case of LiODOM and T-LOAM we took the runtime and environment data from each article respectively. It should be mentioned that, the LeGo-LOAM run data are with out mapping and the ATE and ARE results are 19.57\% and 4.93°/100 m respectively. Our method, by using only 1 core and with out mapping, obtains lower performance results than other similar odometry methods.

\begin{figure}[htbp]
    \centering
    \subfloat[Edge per frames\label{fig:feature_edge}]{
    \includegraphics[width=.95\columnwidth]{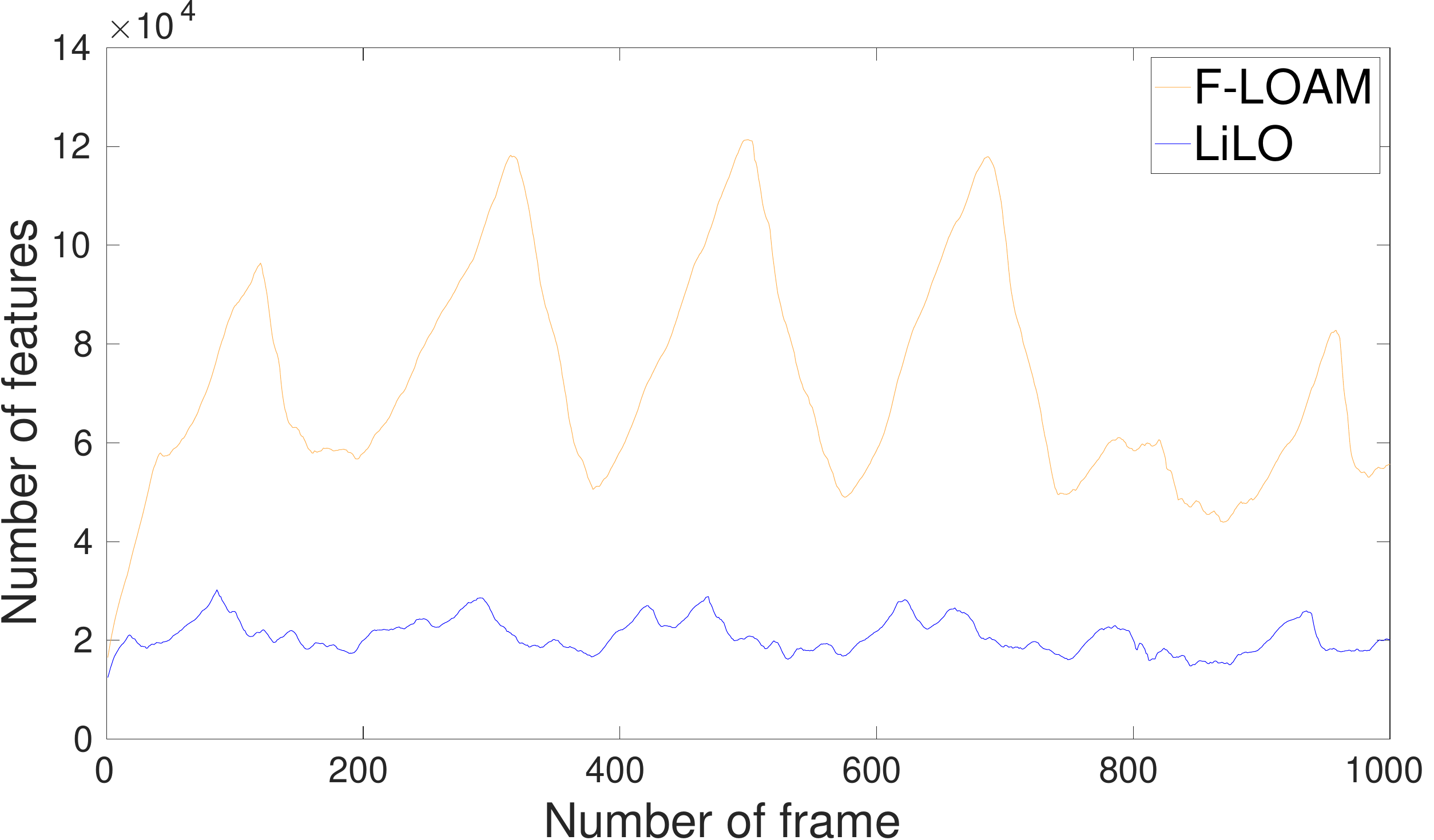}}   
    \\
    \subfloat[ Surf per frames \label{fig:feature_surf}]{
    \includegraphics[width=.95\columnwidth]{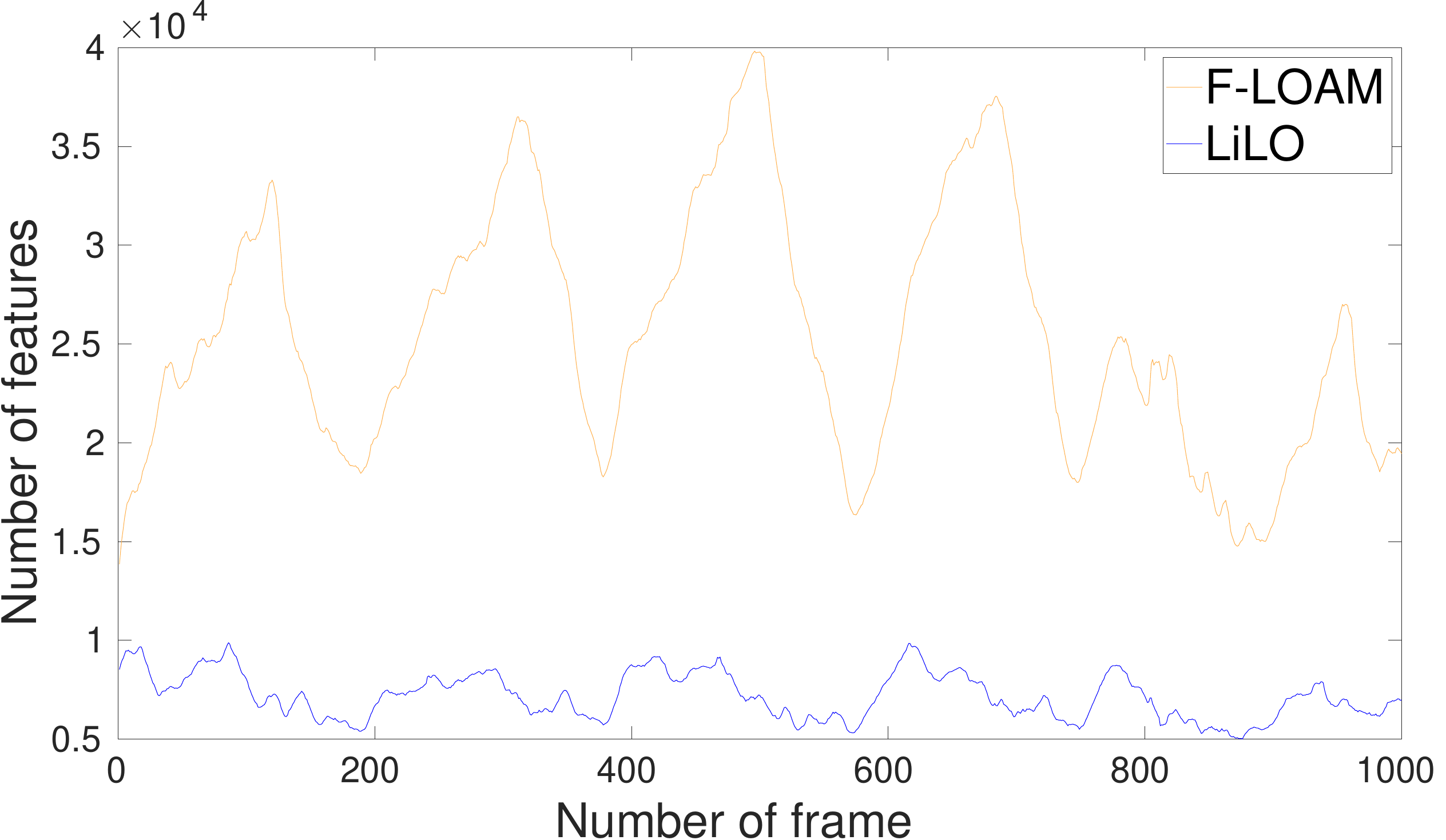}}
    \caption{Comparison of our odometry method and F-LOAM in the number of features required for pose estimation.}

\end{figure}

\begin{figure}[htbp]
    \centering
    \subfloat[Feature extraction time\label{fig:feature_time}]{
    \includegraphics[width=0.95\columnwidth]{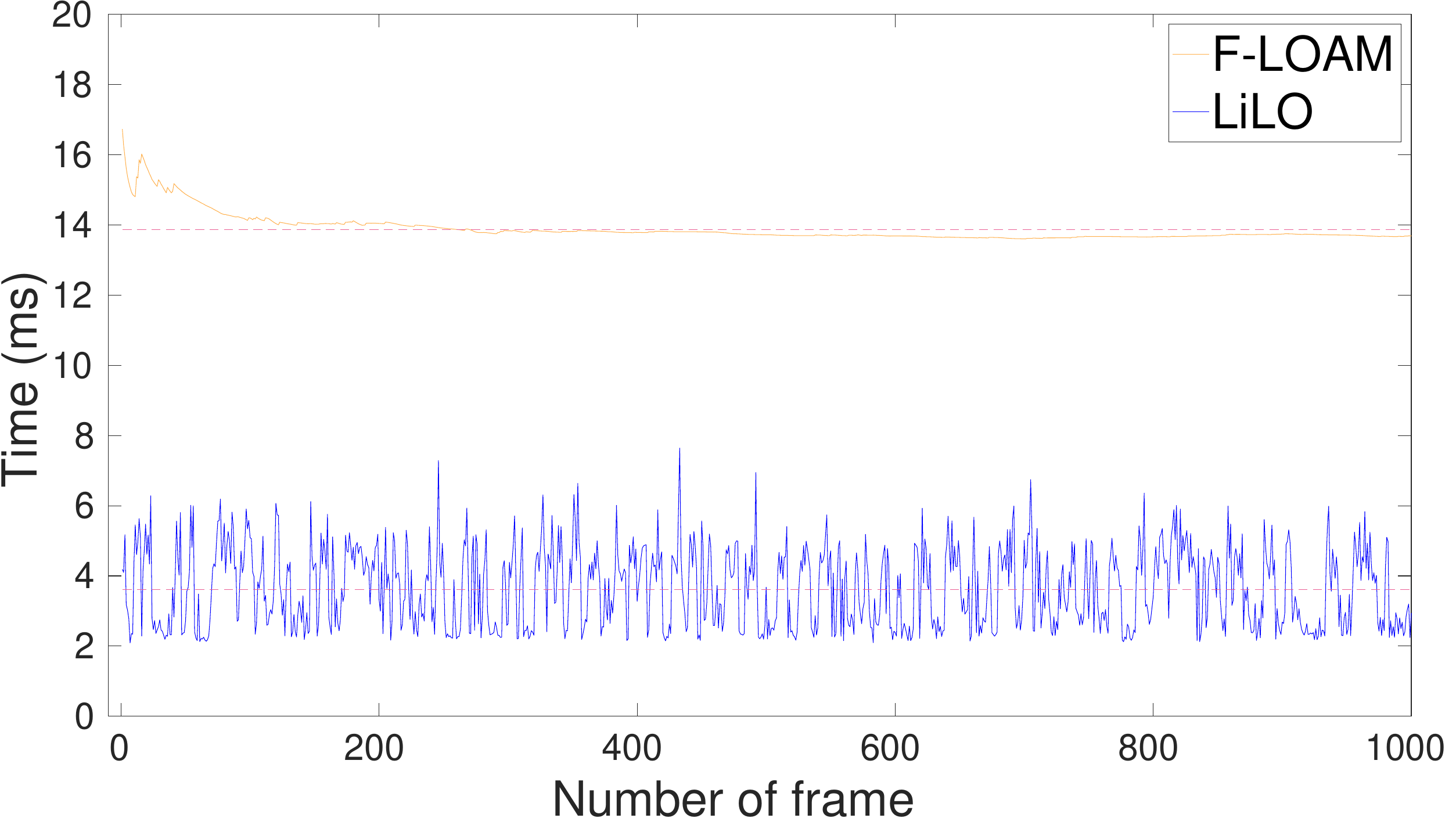}}   
    \\
    \subfloat[Pose estimation time  \label{fig:pose_estimation}]{
    \includegraphics[width=.95\columnwidth]{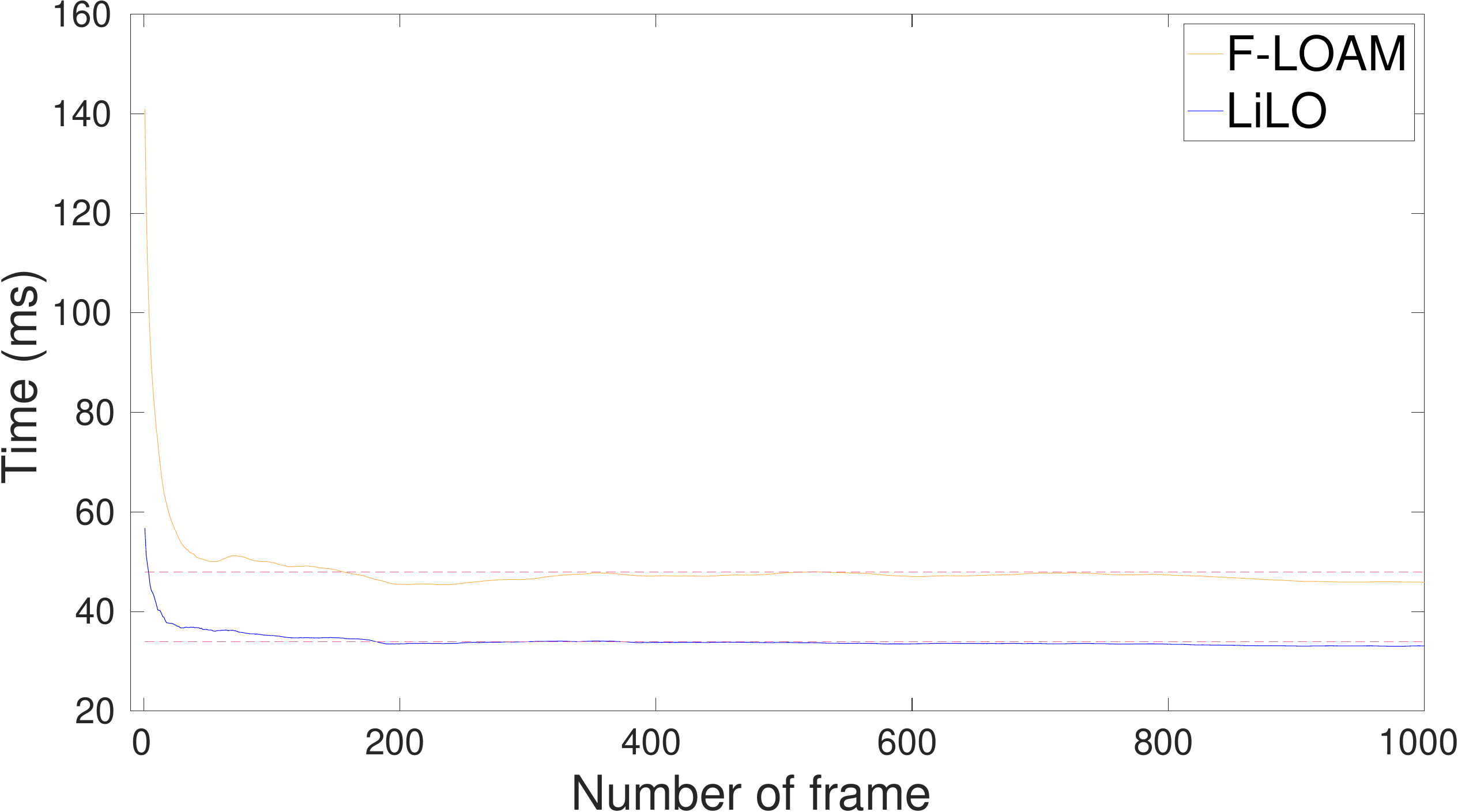}}
    \caption{Runtime of the feature extraction and pose estimation stages of our method compared to F-LOAM.}
\label{fig:runtimeLiLOvsFloam}

\end{figure}

In order to analyze the computational capability of our method we have compared it against F-LOAM on 05 sequence of the KITTI dataset in the same computational environment. Our method shows in Fig. \ref{fig:feature_edge} and \ref{fig:feature_surf} that LiLO uses less surface and edge features respectively than the F-LOAM method for pose estimation. Additionally, Fig \ref{fig:feature_time} and Fig. \ref{fig:pose_estimation} show the point cloud feature extraction and pose estimation time of each method. Our method shows a lower extraction time than F-LOAM as it needs less features to estimate the pose without losing accuracy in the odometry estimation as shown in Table \ref{tab:Kitti table odometry comparation}.

\subsection{Evaluation with our research platform }
\begin{figure}[htbp]
    \centering
    \includegraphics[width=1\columnwidth]{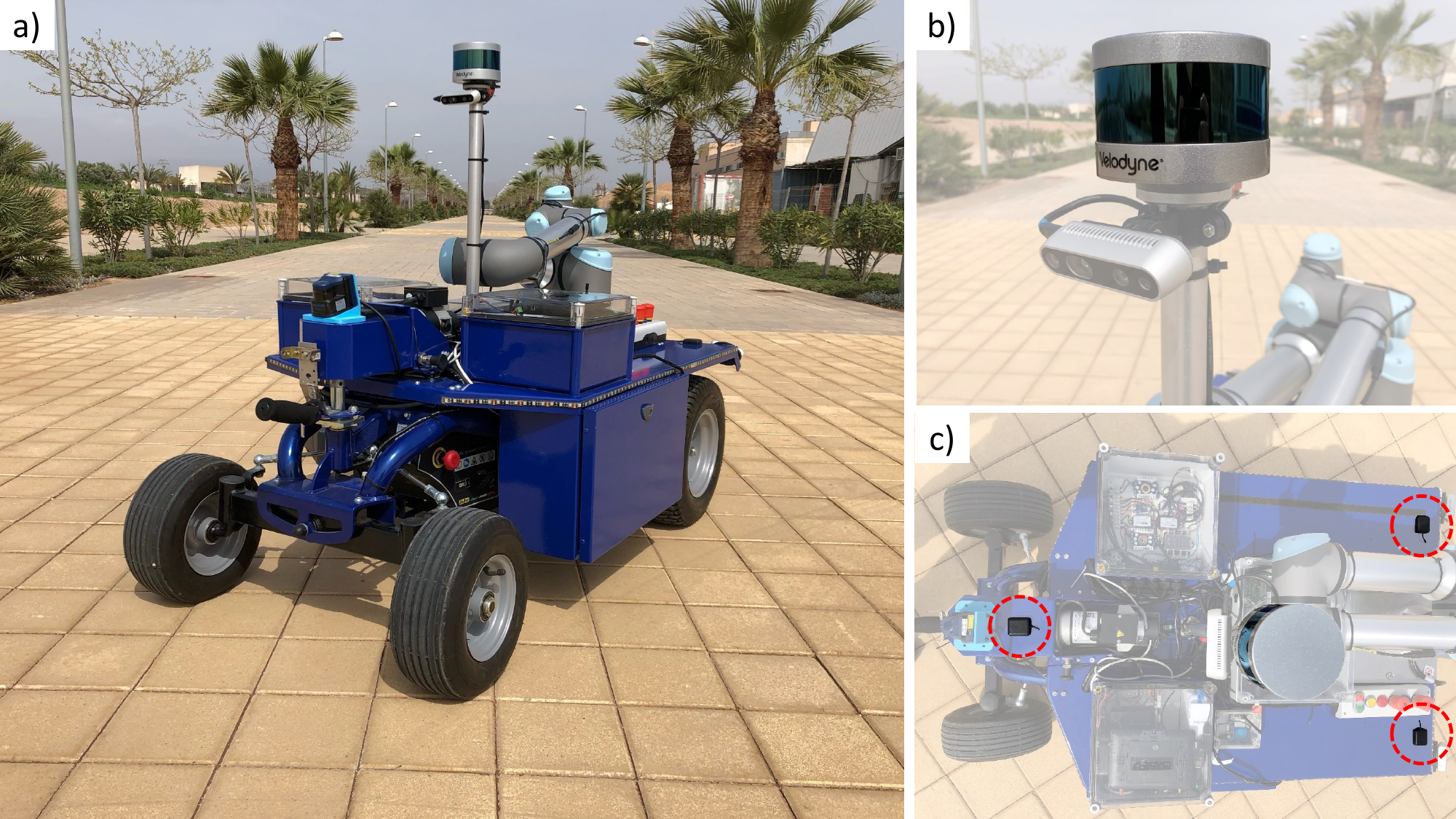}  
    \caption{ ((a) BLUE. roBot for Localization in Unstructured Environments. (b) Velodyne VLP-16 LiDAR sensor. (c) Multi-GNSS system for geolocation.}
    \label{fig:Blue picture}

\end{figure}

We also evaluated the LiDAR odometry estimation presented in this paper on our research platform \textit{BLUE: roBot for Localization in Unstructured Environments} detailed in \cite{BLUE2020deeper} and shown in the Fig. \ref{fig:Blue picture}. This robot has a maximum speed of about 5 km/h and has an Ackermann configuration. It includes an LiDAR 3D Velodyne VLP-16 and a \textit{Multi-GNSS} system with three UbloxNeo-M8N modules. To validate LiLO system, it is compared with F-LOAM, evaluating the error in loop closure in an unstructured outdoor environment. F-LOAM was chosen for the comparison since this method, among the ones analyzed in Section \ref{sec:KITTI dataset evaluation},  is the one that shows the shortest execution time and uses only 1-core for its execution, which allows us to apply it in real time on the mobile robotic platform while having acceptable errors.

\begin{figure}[htbp]
    \centering

    \includegraphics[width=\columnwidth]{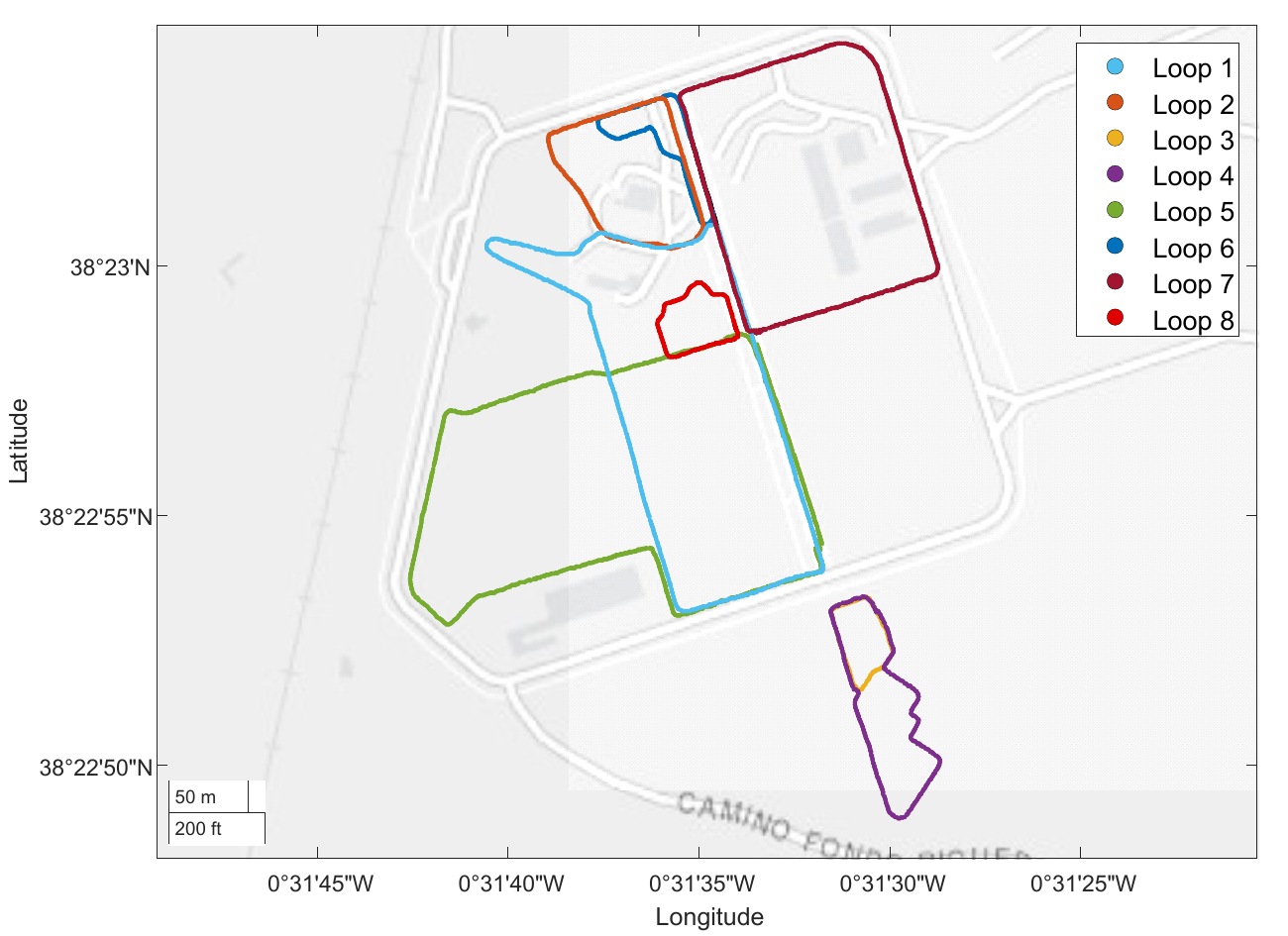}
    \caption{ (Data generated by the Muilti-GNSS system for several paths in the Science Park area at the University of Alicante.}
    \label{fig:truth circuits Gnss}

\end{figure}

In these experiments, we filtered the point cloud image as shown in Section \ref{section:Filtering in the frequency domain}.  The point cloud is converted into a 32x720 resolution image, where the LiDAR data is augmented with 2D interpolation of the data using the bilinear interpolation method shown in \cite{kirkland2010bilinear} and the implementation in \cite{EPVelasco_lidar} . In addition we use a set of \textit{ES} point clouds, eliminating the ground features.
The Velodyne VLP-16 sensor, shown in Fig. \ref{fig:Blue picture}, generates a point cloud with a range of 100 m at a maximum frequency of 20 Hz, supports 16 channels, $\sim$300,000 points per second, a 360° horizontal field and a 30° vertical field of view ($\pm$ 15 up and down). The reference path is obtained with a \textit{Multi-GNSS} system that we implemented, as it detailed in \cite{velasco2021implementacion}. This georeferencing system provides the longitude and latitude data at 3 Hz with three \textit{UbloxNeo-M8N}. 

\begin{table*}
\caption{Comparative experiments of LiLO and F-LOAM. Closure errors (m) of the loops in Fig. \ref{fig:truth circuits Gnss}. The value \textbf{d} is the modulus of the initial and final  position of each experiment. Runtime (ms) is the odometry estimation time per frame.}
    \centering
    \small
    \begin{tabular}{llllllllllll}
    \toprule
    & \textbf{Loop}             &  & \textbf{1}& \textbf{2}& \textbf{3}& \textbf{4}& \textbf{5}& \textbf{6}& \textbf{7}& \textbf{8}&\\
    & \textbf{Distance (m)}     &  & 767& 151  & 351& 900& 580& 164& 278& 318& \multicolumn{1}{l}{\textbf{\textbf{Average}}}  \\ 
    \hline
    & \textbf{Runtime}&& 27.8& \textbf{19.0}& \textbf{28.8}& 28.1& \textbf{24.9}& 29.1& 29.1& 28.0& 26.9\\
    && \textbf{x}& 0.26& 0.40& -0.06& 0.64& -0.31& -0.07& 0.04& -0.06& -\\
    \textbf{\textbf{LiLO}}& \textbf{\textbf{Closure}}& \textbf{y}& 0.85& 0.13& 0.31& 0.97& 0.62& -0.01& -0.12& -0.04& -\\
    & \textbf{Error (m)} & \textbf{z}& 0.15& 0.02& -0.19& -3.28& -0.69& -0.06& 0.26& 0.06& -\\
    && \textbf{d}& \textbf{0.90}& \textbf{0.42}& \textbf{0.37}& \textbf{3.48}& \textbf{0.98}& 0.10& \textbf{0.29}& \textbf{0.09}& \textbf{0.83}\\ 
    \hline
    &  \textbf{\textbf{Runtime}}&& \textbf{27.1}& 20.1& 30.2& \textbf{27.5}& 25.8& \textbf{28.0}& \textbf{28.0}& \textbf{27.9}& \textbf{26.8}\\
    && \textbf{x}& 0.17& 0.31& 0.77& -2.00& 0.29& -0.07& 0.05& -0.17& -\\
    \textbf{\textbf{F-LOAM}} &  \textbf{\textbf{Closure}}& \textbf{y}& 2.07& 0.27& 0.14& 2.23& 1.50& -0.02& -0.13& -0.02& -\\
    &  \textbf{Error (m)} & \textbf{z}& -3.03& 1.06& -7.83& -5.08& -7.35& 0.00& 0.30& -0.04& -\\
    && \textbf{d}& 3.67& 1.14& 7.87& 5.90& 7.50& \textbf{0.07}& 0.33& 0.18& 3.33\\ 
    \hline
    \end{tabular}

\label{tab:Scientific_Park table}
\end{table*}

\begin{figure*}[htbp]
\centering

     \subfloat[Loop 1 \label{fig:fig_Loop_00}]{
      \includegraphics[width=0.24\textwidth]{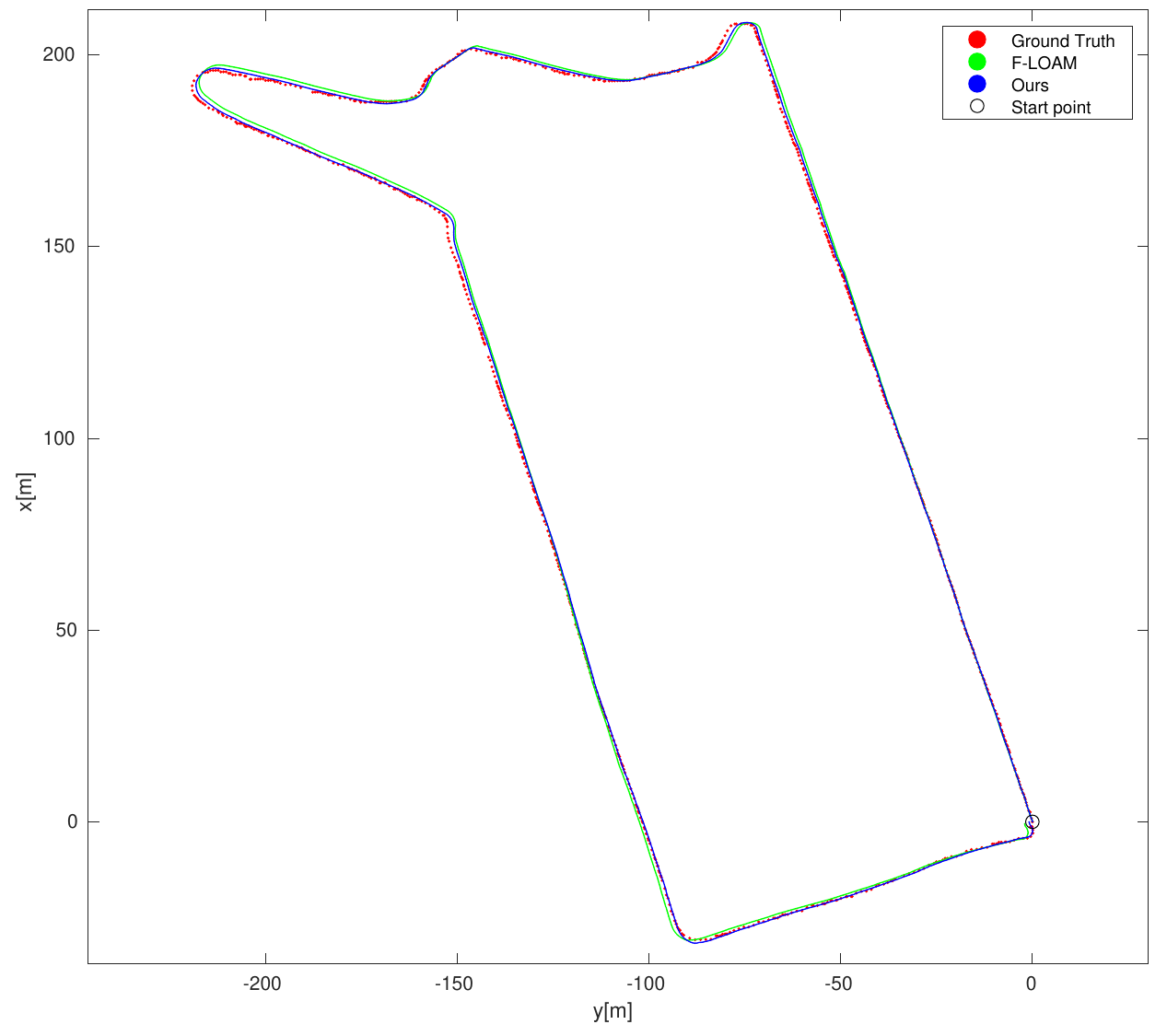}}
      \hfill
      \subfloat[Loop 2 \label{fig:fig_Loop_01}]{
      \includegraphics[width=0.24\textwidth]{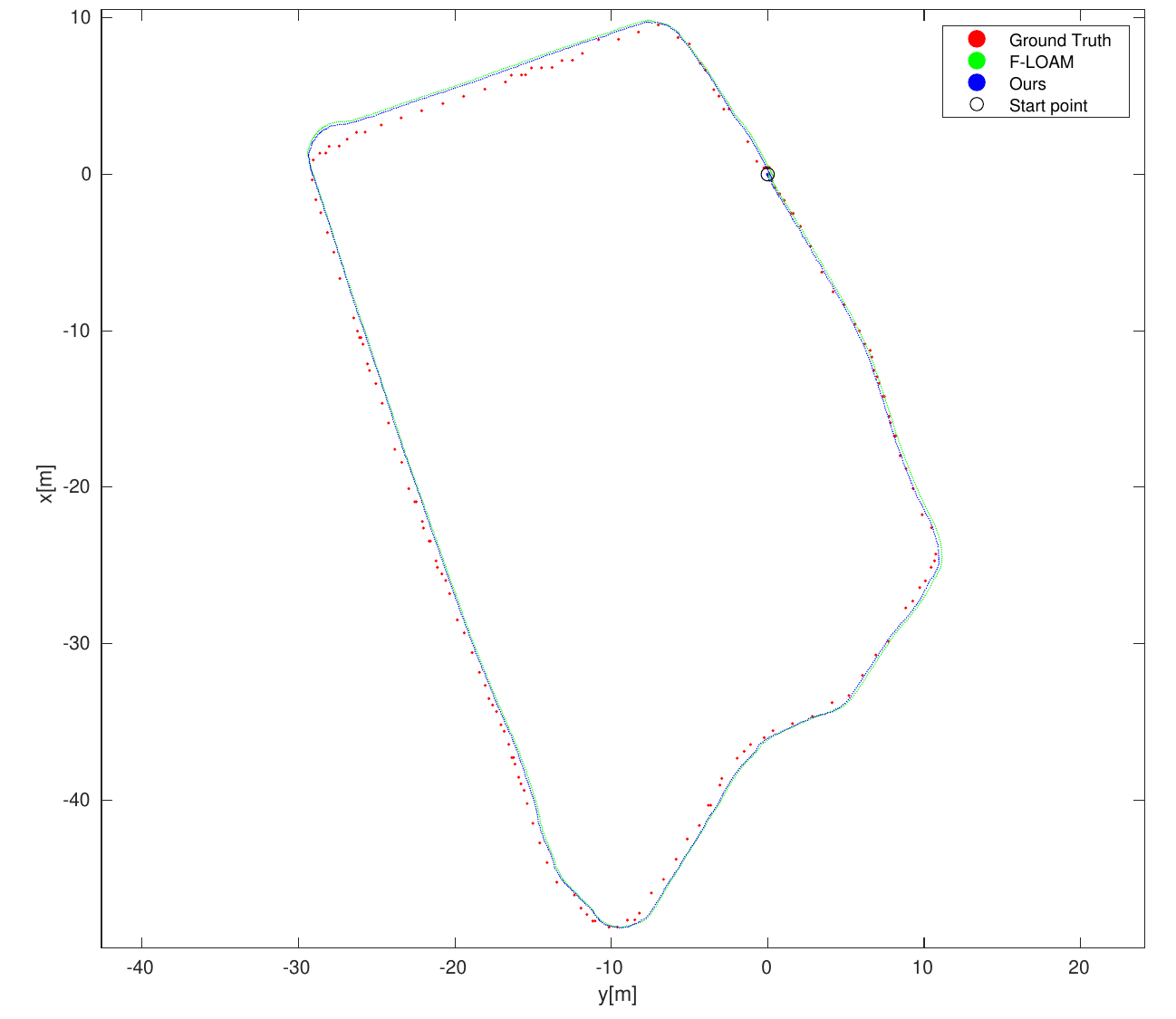}}
      \hfill
      \subfloat[Loop 3 \label{fig:fig_Loop_02}]{
      \includegraphics[width=0.24\textwidth]{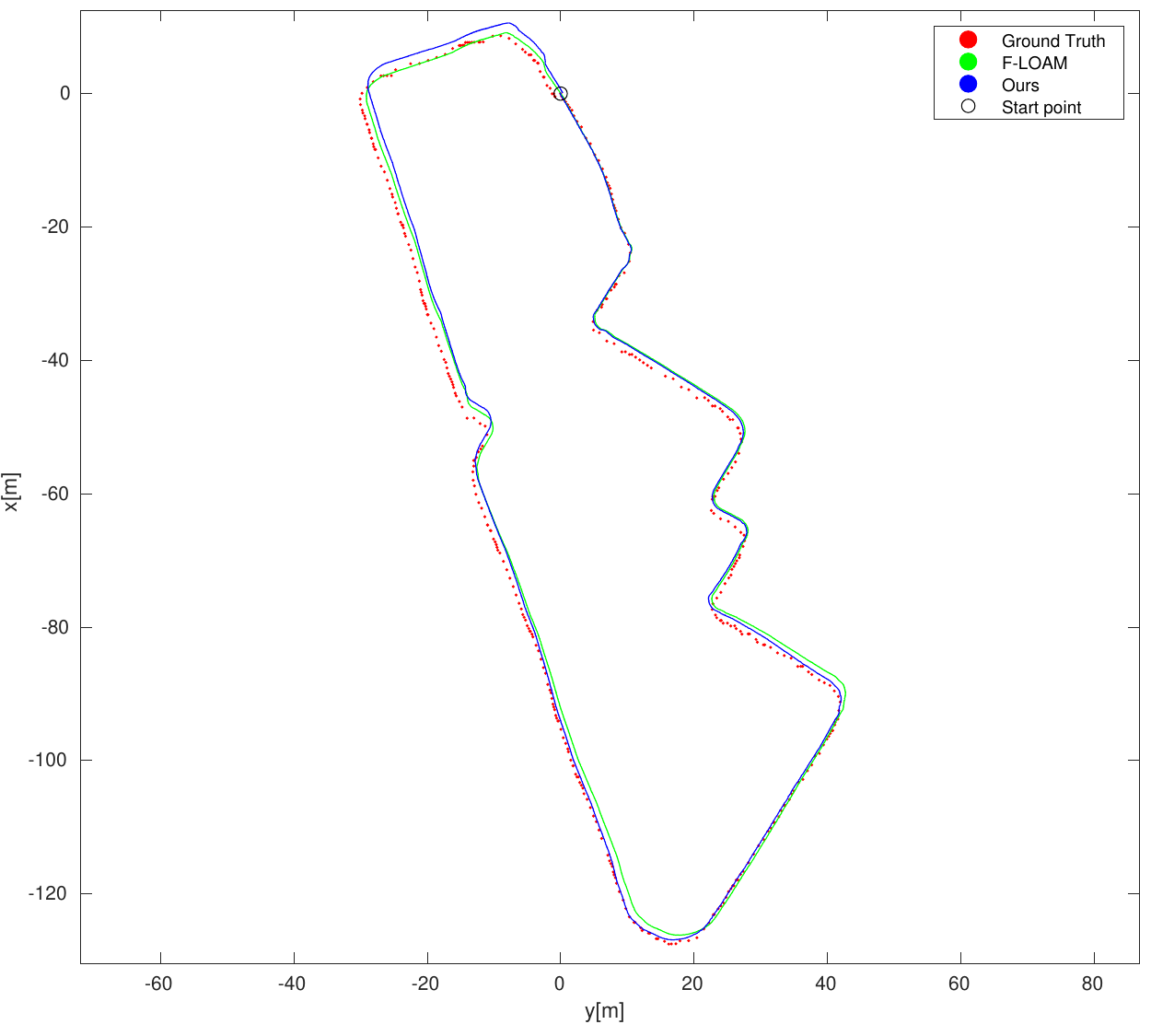}}
      \hfill
      \subfloat[Loop 4 \label{fig:fig_Loop_03}]{
      \includegraphics[width=0.24\textwidth]{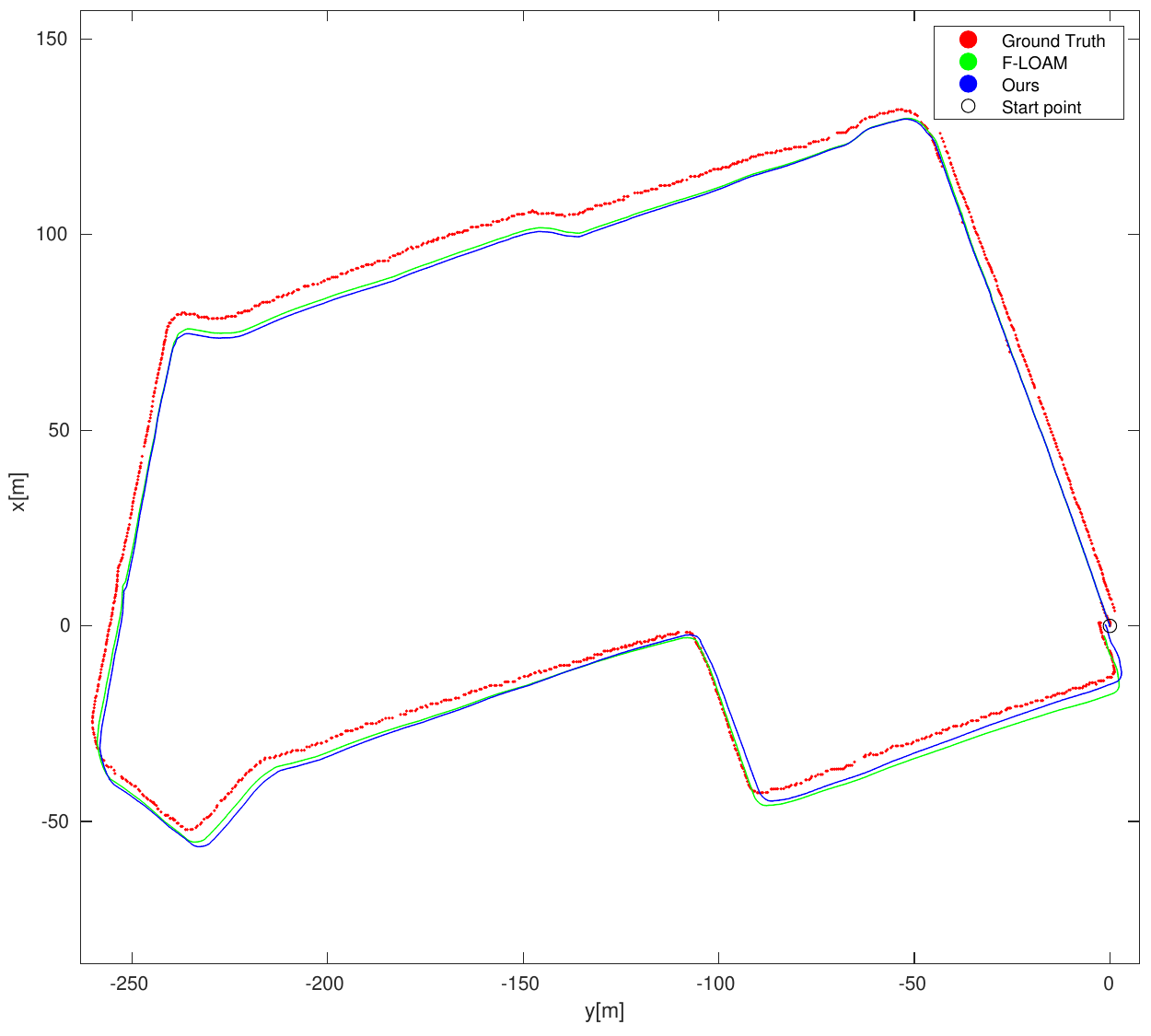}}
      \\
      \subfloat[Loop 5 \label{fig:fig_Loop_04}]{
      \includegraphics[width=0.24\textwidth]{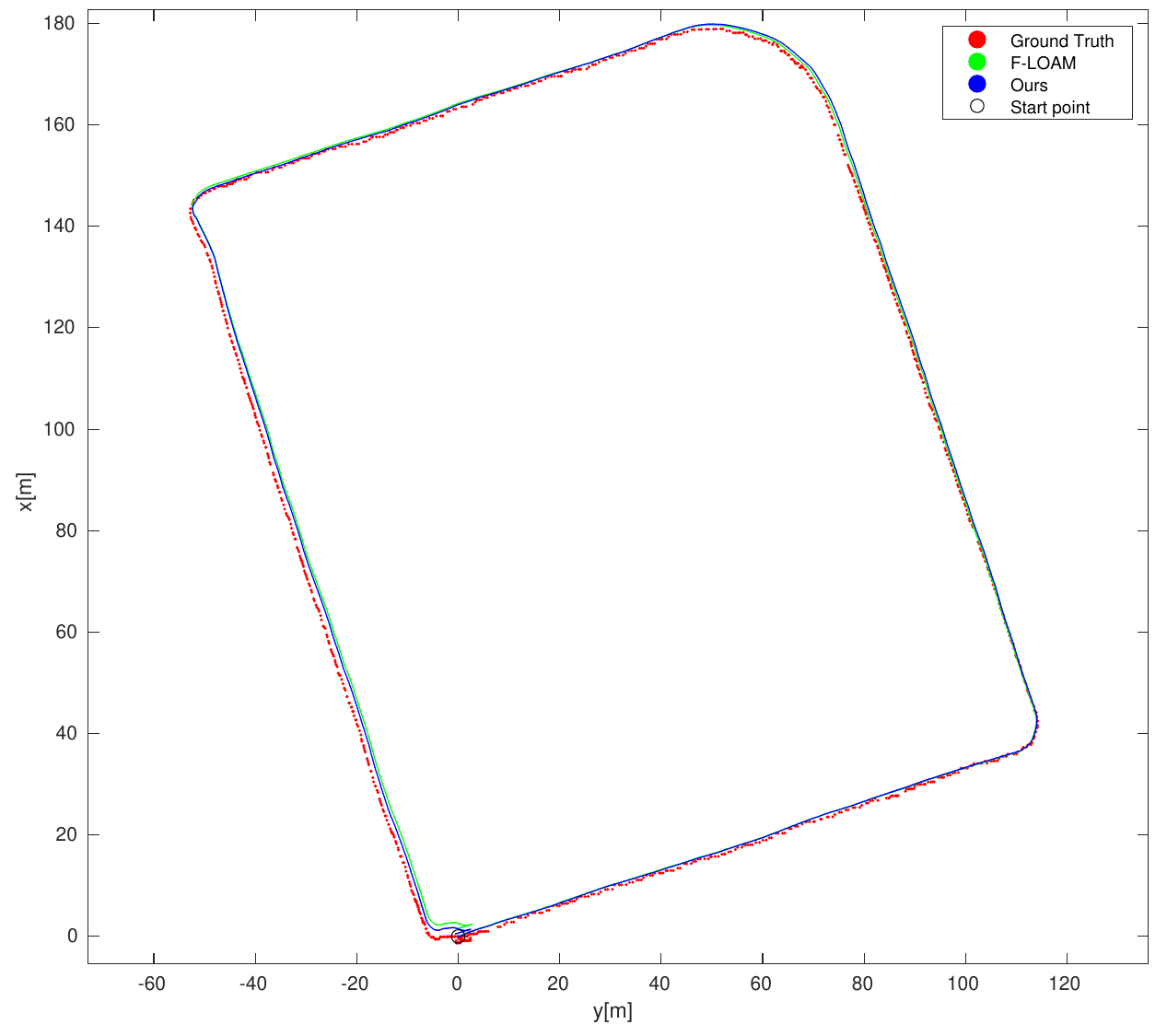}}
      \hfill
      \subfloat[Loop 6 \label{fig:fig_Loop_05}]{
      \includegraphics[width=0.24\textwidth]{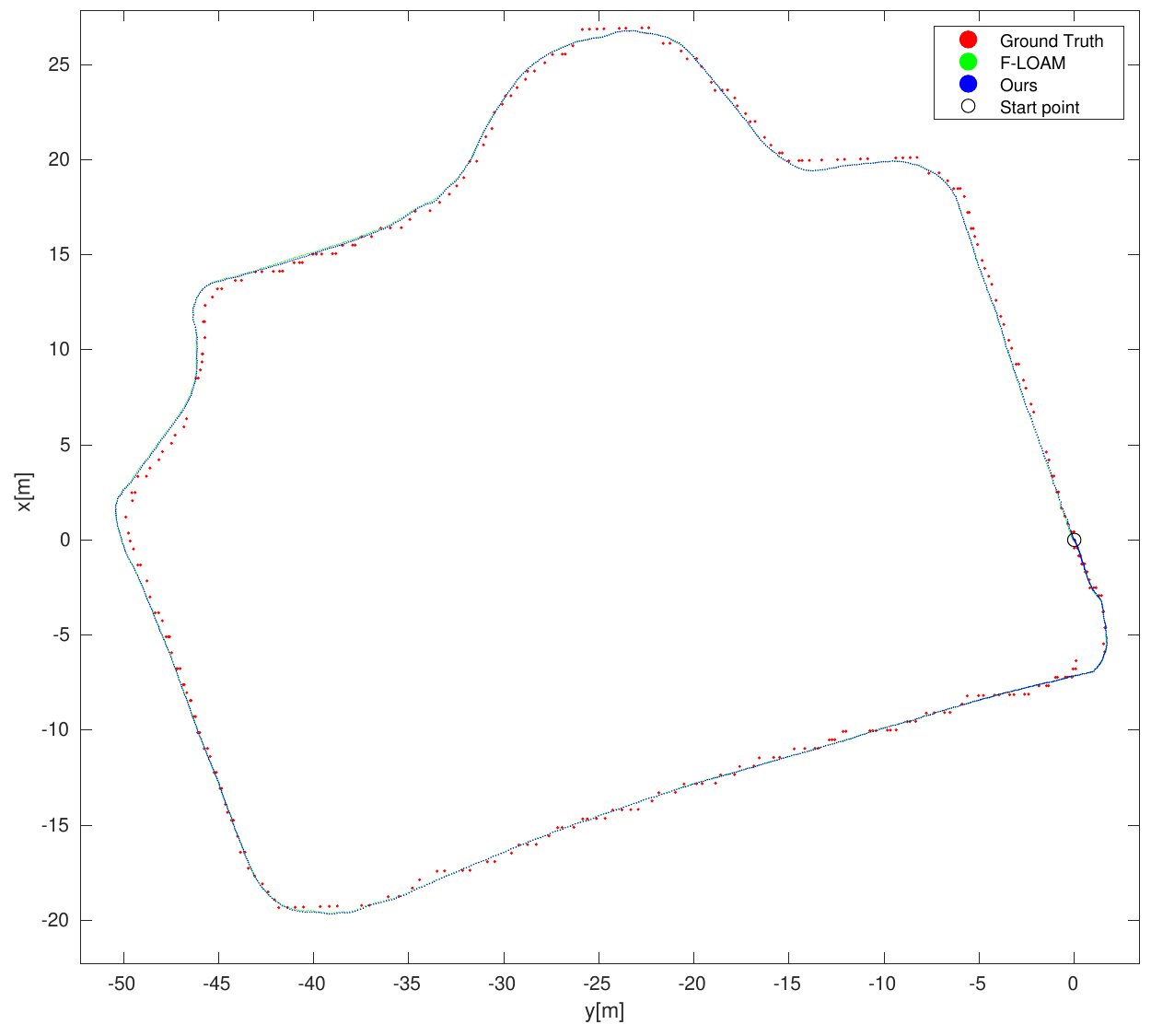}}
      \hfill
      \subfloat[Loop 7 \label{fig:fig_Loop_06}]{
      \includegraphics[width=0.24\textwidth]{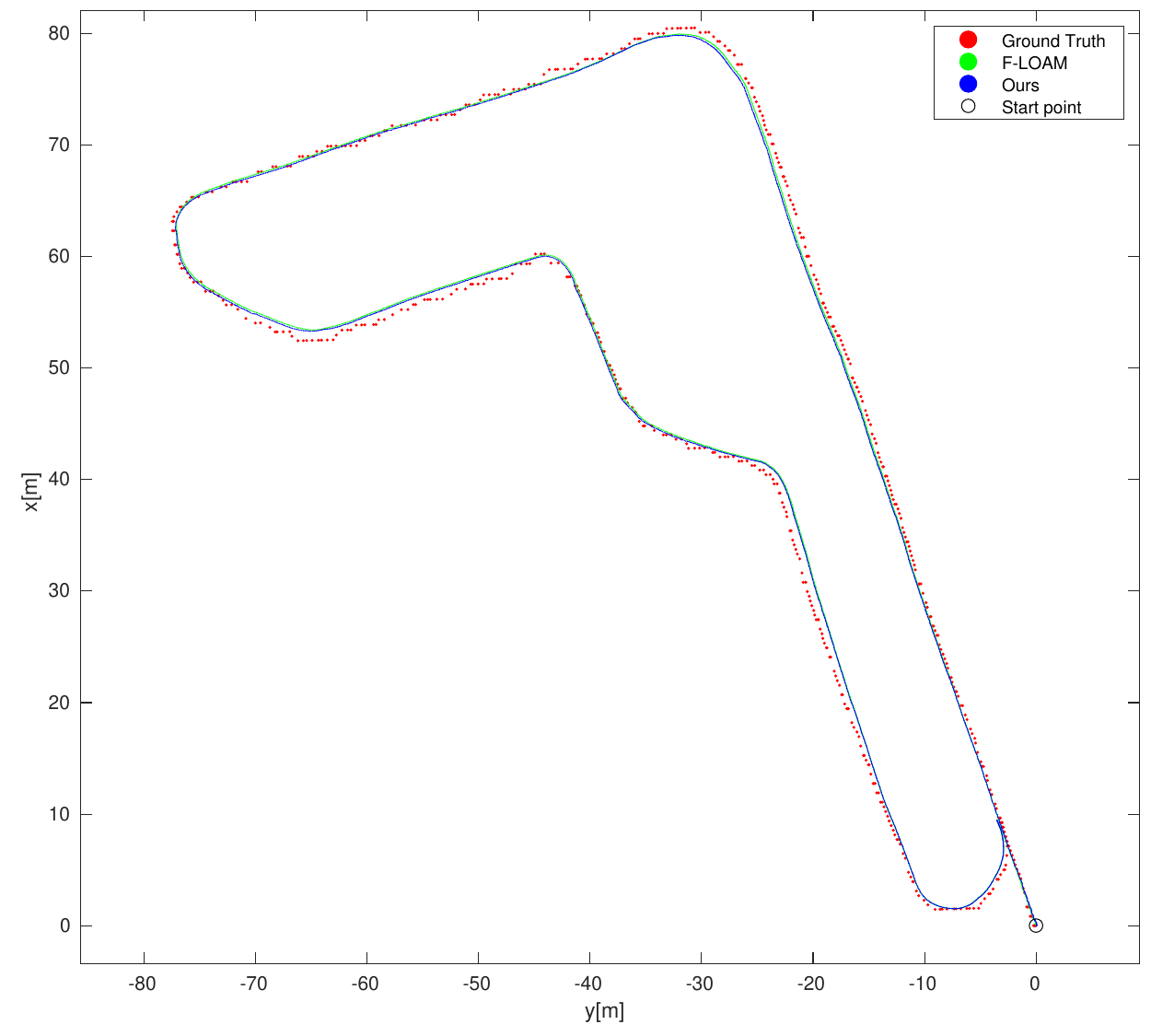}}
      \hfill
      \subfloat[Loop 8 \label{fig:fig_Loop_07}]{
      \includegraphics[width=0.24\textwidth]{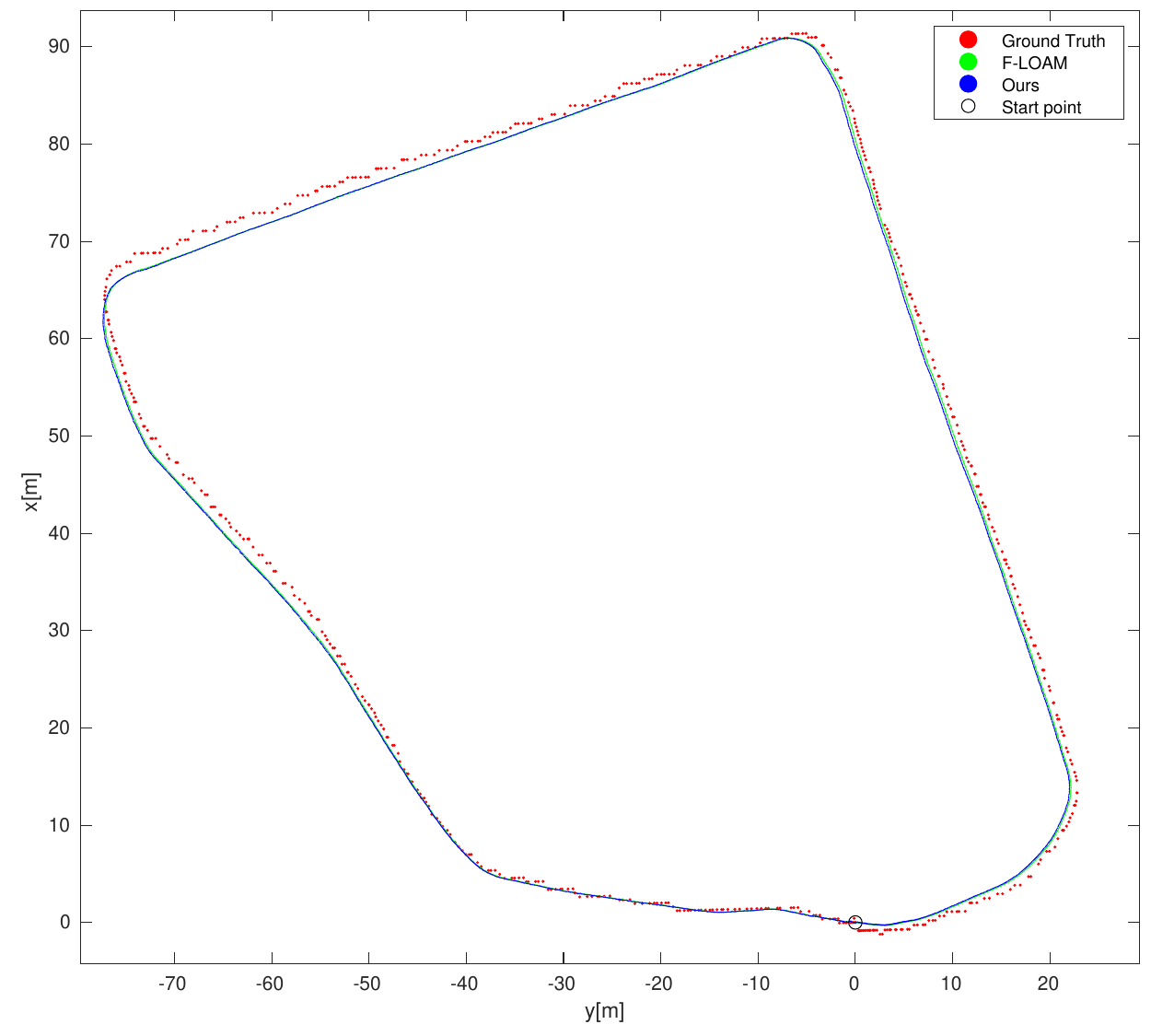}}

\caption{
Results of applying our odometry method (blue) to estimate the pose on the BLUE platform, in comparison with the estimation given by the F-LOAM method (green), and the reference path of the \textit{Multi-GNSS} georeference system (red), for the eight real paths.}
\label{fig:all_loops}
\end{figure*}

Fig. \ref{fig:truth circuits Gnss} shows the reference paths of the eight loops covering a total of 3509 m, for which we created a point cloud database from the Velodyne VLP-16  LiDAR 3D sensor and Multi-GNSS georeferences. The Multi-GNSS system was used as a visual reference for experiments and does not represent the ground truth due to georeferencing errors in the system. For this reason, the \textit{UTM} (Universal Transverse Mercator) coordinates shown in Fig. \ref{fig:all_loops} are not used to calculate the rotation and translation errors in each measurement, as shown in experiments in Section \ref{sec:KITTI dataset evaluation}. These georeferencing data were converted to \textit{UTM} using the \textit{deg2utm} function developed in \cite{degutmMatlab}. In each loop, the initial point $\{x_0, y_0, z_0\}$ is defined as the first latitude and longitude value converted to \textit{UTM}. This transformation has only Cartesian coordinates in the ground plane as $\{x, y\}$, therefore, the current path of these loops has no $z$ data. Hence, we considered the same value of $z$ along the path. 

\begin{figure*}[htbp]
    \centering
    \includegraphics[width=0.7\textwidth]{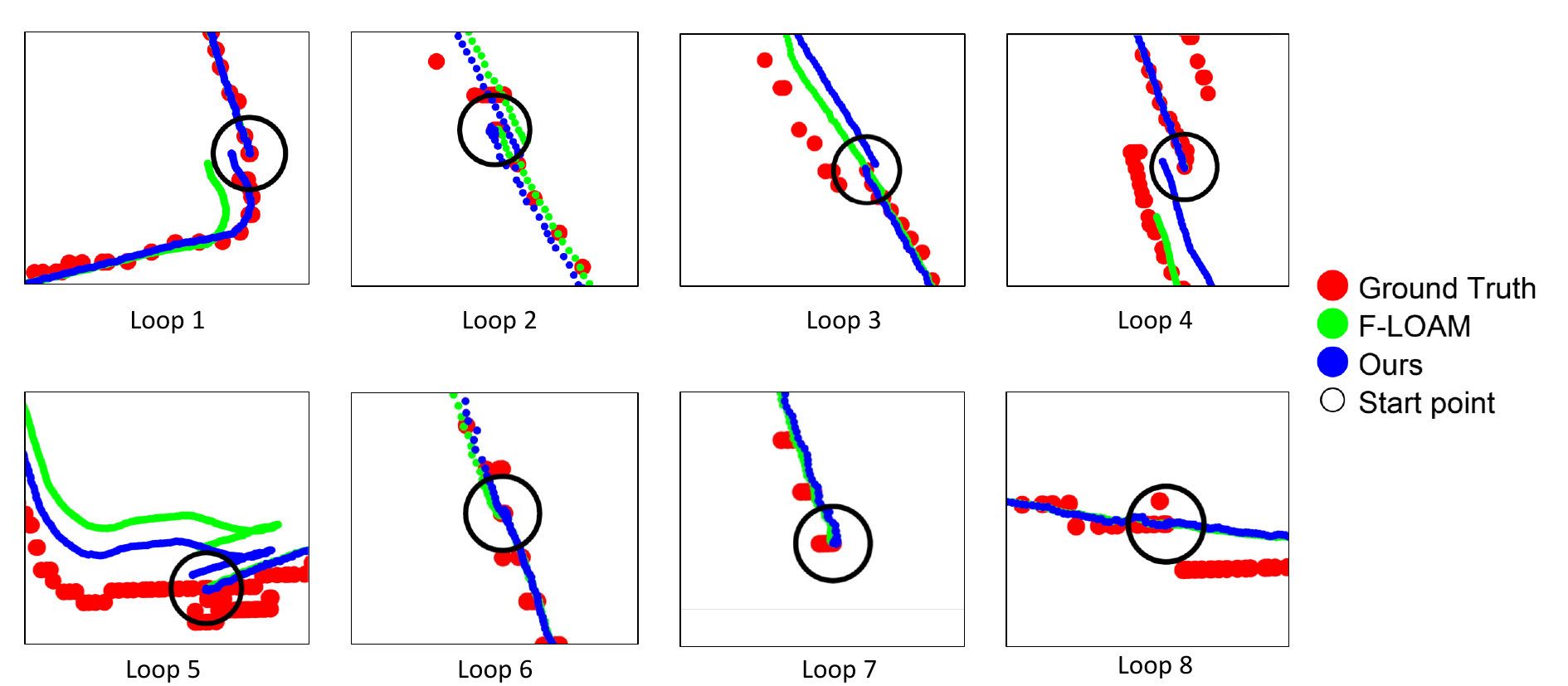}
    \caption{Loop closure errors of loops number 1, 3 , 7. The green trajectory is generated by F-LOAM, the blue trajectory is generated by our LiDAR odometry system and the red points are the reference positions of the Multi-GNSS system.}
    \label{fig:loop_closure_errors}
\end{figure*}

Fig. \ref{fig:loop_closure_errors} shows the loop closure comparison of loops 1, 3 and 7 compared to the F-LOAM method and our method. LiLO achieves better loop closure and approximates the geo-reference of the Multi GNSS system, showing that our method has a lower bias.

For this reason, to compare the quality of LiLO system, the error is determined by closing the loop, where the start and end points should be the same. The results in table \ref{tab:Scientific_Park table} compare errors of LiLO and F-LOAM methods, where our method has a loop closure error 4 times lower than the F-LOAM with a similar execution time. As demonstrated in section \ref{fig:runtimeLiLOvsFloam}, LiLO is faster than F-LOAM. In this experiment, the execution time of LiLO is similar to F-LOAM because it worked with a spherical image range of 32x720, i.e., the LiDAR sensor layers are virtually increased from 16 to 32 layers. In this way, we have increased the size of data to be processed reducing the loop closure error with a similar execution time with the method compared to F-LOAM.

\section{Conclusion}
\label{sec:cloncusion}
In this paper, we propose the LiLO method, which is a a lightweight, low-bias and computationally efficient LiDAR odometry method that is based on feature detection in a SRI of the point cloud and LOAM-based pose estimation methods. The experiments performed show a significant improvement in translation and rotation errors with a reduced runtime compared to LOAM-based techniques. This reduction in runtime represents a computationally lighter system.  In addition, in experiments with a 16 laser beam LiDAR sensor, a significant improvement in loop closure errors without the need for global maps is observed compared to F-LOAM. Finally, LiLO presents an improvement in the execution time and the number of features needed for pose estimation. Therefore, our odometry approach with feature filtering on SRI is lightweight and has low-bias which can be used as a stepping stone to complete localization systems. Given the good results of the proposed method, we continue working on its improvement and application in real mobile robots.


\subsubsection*{Acknowledgements.} This work has been supported in part by the Spanish Government through the research project PID2021-122685OB-l00, as well as the grants for Training of Research Staff PRE2019-088069, from the Government of Spain, and ACIF/2019/088 from the Regional Valencian Community Government and the European Development Fund.

\subsubsection*{Author Contributions.} All authors contributed to the conception and design of the study. Material preparation, data collection, and analysis were performed by E.V and M.M. The first draft of the manuscript was written by E.V. and all authors commented on previous versions of the manuscript. The illustrations were done by E.V. All authors read and approved the final manuscript.

\subsubsection*{Funding.} This work is funded by the Spanish Ministry of Science and Innovation, the Generalitat Valenciana and the European Development Fund.

\subsection*{Declarations.}
\textbf{Ethical statement} The authors have no competing interests to declare that are relevant to the content of this article. 

\bibliography{reference}

\end{document}